\documentclass[preprint, authoryear]{elsarticle}
\usepackage{graphicx}
\usepackage[colorlinks = true,
            linkcolor = blue,
            urlcolor  = blue,
            citecolor = blue,
            anchorcolor = blue]{hyperref}
\usepackage{multirow}
\usepackage[normalem]{ulem}
\useunder{\uline}{\ul}{}
\usepackage{bm}
\usepackage{booktabs}
\usepackage{float}
\usepackage{amsmath}
\usepackage{amssymb}
\usepackage{caption}
\usepackage{array}
\usepackage[table,xcdraw]{xcolor}

\journal{a journal}

\begin{document}
\captionsetup[figure]{labelfont={bf},labelformat={default},labelsep=period,name={Fig.}}
\captionsetup[table]{labelfont={bf},labelformat={default},labelsep=newline,name={Table},singlelinecheck=off}
\begin{frontmatter}
\title{Daily peak electrical load forecasting with a multi-resolution approach}


\author[mymainaddress,mysecondaryaddress]{Yvenn Amara-Ouali}\corref{mycorrespondingauthor}
\author[thirdaddress]{Matteo Fasiolo}
\author[mymainaddress,fourthaddress]{Yannig Goude}
\author[fourthaddress]{Hui Yan}

\address[mymainaddress]{Laboratoire de Mathématiques d’Orsay (LMO), CNRS, Université
Paris-Saclay, Faculté des Sciences d'Orsay, bat 307, 91405 Orsay,
France}
\address[mysecondaryaddress]{CELESTE, Inria Saclay, FRANCE}
\address[thirdaddress]{School of Mathematics, University of Bristol, Bristol, UK}
\address[fourthaddress]{EDF Lab, 7 bd Gaspard Monge, 91120 Palaiseau, France}

\begin{abstract}

In the context of smart grids and load balancing, daily peak load forecasting has become a critical activity for stakeholders of the energy industry. An understanding of peak magnitude and timing is paramount for the implementation of smart grid strategies such as peak shaving. The modelling approach proposed in this paper leverages high-resolution and low-resolution information to forecast daily peak demand size and timing. The resulting multi-resolution modelling framework can be adapted to different model classes. The key contributions of this paper are a) a general and formal introduction to the multi-resolution modelling approach, b) a discussion on modelling approaches at different resolutions implemented via Generalised Additive Models and Neural Networks and c) experimental results on real data from the UK electricity market. The results confirm that the predictive performance of the proposed modelling approach is competitive with that of low- and high-resolution alternatives.
\end{abstract}


\end{frontmatter}

\section{Introduction}

The electric daily peak load is the maximum of the electricity power demand curve over one day. Having an accurate forecast of the daily peak enables independent system operators (ISOs) and energy providers to better deliver electricity and optimise power plant schedules. The importance of such a forecast is increasing as the integration of intermittent renewable production sources progresses. In particular, renewable energy sources are at the bottom of the merit order curve which makes them (currently) the most economical source of energy used to serve the market. However, they are intermittent and provide time-varying levels of power generation, which are only partially under human control. If electricity demand is high and renewables cannot provide for it alone, ISOs have to deliver electricity from sources with higher marginal costs (e.g., gas-fired plants) for the stakeholders as well as for the environment in terms of CO2 emissions. In such a context, accurately forecasting the peak demand magnitude and timing is essential for determining the generation capacity that must be held in reserve. 

Electrical equipment is tailored to support a specific peak load. If the demand comes close or exceeds the network capacity, it can lead to distribution inefficiencies and ultimately power system failures, such as blackouts. With the increasing number of electric vehicles (EV) in circulation, a further source of stress is added to the electricity system. For instance, 46\% of vehicles sold in Norway in 2019 were EVs \citep*{international_energy_agency_global_2019}. The challenge posed by the additional EV demand must be met by more tailored management systems and policies, if expensive infrastructural works are to be avoided. Dynamic electricity pricing schemes, for example, the Triads in the UK or the Global Adjustment in Ontario, Canada, have been developed to reduce the system peak load. Consumers who can correctly estimate and cut their use during peak events can unlock great savings. Peak demand forecasts will thus be key for the development of such policies.

\begin{figure}[H]
    \centering
    \includegraphics[width=\linewidth,keepaspectratio]{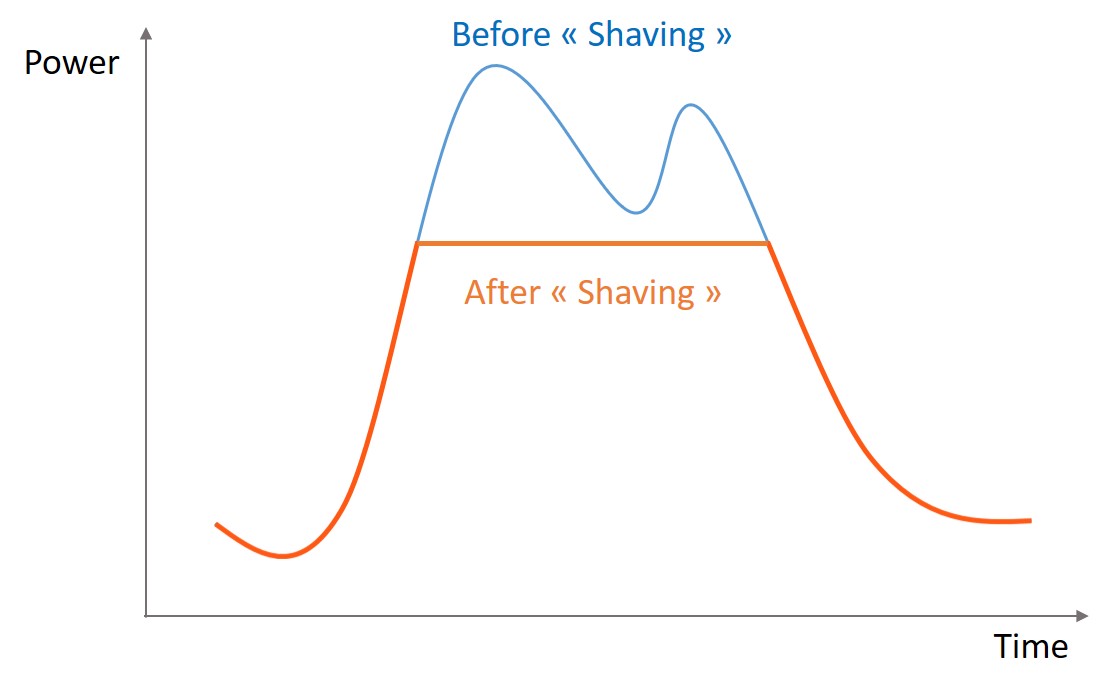}
    \caption{Illustration of peak shaving}
    \label{Shaving}
\end{figure}

To account for the increasing demand for electricity and to prevent system failures, smart grid technologies and policies are being implemented to foster communication between the various stakeholders of the electricity supply chain to achieve a more efficient use of energy. One major objective is to maximise the load factor. The load factor is the average load over a specific time period divided by the peak load over the same period. Maximising it leads to a more even use of energy through time, thus preventing system failures and surges in electricity prices. One of the most common ways to achieve load factor maximisation is peak shaving (Figure \ref{Shaving}), which refers to the flattening of electrical load peaks. Three major strategies have been proposed for peak shaving, namely integration of Energy Storage System (ESS), integration of Vehicle-to-Grid (V2G) and Demand Side Management (DSM) \citep*{uddin_review_2018}. ESS and V2G integration provide ancillary sources to balance the grid through batteries while DSM shifts consumer demand to flatten the peak. To be activated adequately, all these strategies require accurate forecasts of the demand peak magnitude (DP) and of the instant at which it occurs (IP).

This article proposes novel methods to forecast the DP and the IP by leveraging information at different time resolutions. In particular, the multi-resolution approach proposed here is illustrated in the context of two model classes: Generalised Additive Models (GAMs) and Neural Networks (NNs). Both are state of the art predictive models, widely used to forecast electrical load in industry and academia. The performance of the multi-resolution framework under both model classes is assessed using aggregate UK electricity demand data from the National Grid \citep*{nationalgrid_eso_2021}.

The rest of the paper is structured as follows: Section 2 presents a literature review of daily peak forecasting methodologies. Section 3 introduces multi-resolution modelling using GAMs and neural networks. Section 4 explains how the different models were set up in the high-resolution, low-resolution and multi-resolution settings. Section 5 analyses the results of the models described in Section 4, using UK demand data.

\section{Related work}

This section provides an extensive literature review of peak forecasting methods and was conducted to identify gaps in the field. It includes methods ranging from probabilistic approaches to deep learning.

Probabilistic forecasts have been widely adopted in the context of load forecasting applications \citep*[e.g.,][for an overview]{hong_probabilistic_2016}, but little has been done on probabilistic peak demand forecasting. Two probabilistic set-ups, commonly used for peak load forecasting, were outlined by \cite*{jacob_forecasting_2020}. The first is block maxima (BM), where data is separated into time chunks of equal lengths and the maximum of each chunk is assumed to approximately follow a generalised extreme value (GEV) distribution. The second is peaks over threshold (POT), which approximates the distribution of the excess load over a threshold by a generalised Pareto distribution. While the POT and BM settings can be unified via point processes \citep*{boano-danquah_analysis_2020}, in this work we are mainly interested in the BM case.

In a long-term forecasting setting, \cite*{mcsharry_probabilistic_2005} used demand data at the daily resolution to forecast the magnitude and timing of the yearly peak (i.e., the day characterised by the largest total demand). They considered a forecasting lead time of one full year and obtained a probabilistic forecast by simulating year-long trajectories for the weather variables and plugging them into a deterministic linear regression model. Similarly, \cite*{hyndman_density_2010} considered a long-term forecasting application, where the aim was to forecast the probability distribution of the annual and weekly peak electricity demand. They used semi-parametric additive models to capture the effect of covariates, such as temperature, on the demand and obtained a probabilistic forecast by adopting a simulation and scenario-based approach. \cite*{elamin_quantile_2018} used quantile regression methods to forecast the DP one day ahead. Even though they used quantile regression to obtain an upper bound on demand, quantile estimates at several probability levels could be used to estimate the full peak demand distribution. Also \cite*{gibbons_quantile_2014} modelled the DP via a quantile regression model, but their objective was post-processing daily estimates to forecast the annual demand peak, rather than modelling the DP probabilistically.

Multivariate regression models using multivariate adaptive regression splines (MARS) were proposed by \cite*{sigauke_daily_2010} to forecast the DP in South Africa. Explanatory variables including meteorological variables are aggregated at the daily resolution (e.g., average, minimum and maximum temperature). The model outperforms piecewise polynomial regression models with an autoregressive error term. \cite*{sigauke_prediction_2011} studied time series of the DP and illustrated its heteroscedastic structure. A SARIMA–GARCH errors model and a regression-SARIMA–GARCH model are then proposed to forecast it at a short-term horizon. Results show that SARIMA-like models produce forecasts with an accuracy around 1.4 in mean absolute percentage error on a testing period. 

\cite*{saxena_hybrid_2019} proposed a hybrid model to forecast whether the following day will be a peak load day for the billing period for customers subject to demand charge structure. They apply their model to optimise the electricity bill of an American University. Load data is provided every five minutes from January 2013 to April 2016. Here, the POT set-up was used with a threshold depending on a monthly average and variance of the daily load. An original combination of 4 forecasts was proposed. First, a linear model is used to forecast the maximum daily load at a monthly horizon which is then coupled to short-term load forecasting models (NN and ARIMA) to provide two forecasts. Two other forecasts were computed using binary classifiers (logistic regression and NN) and a synthetic minority over-sampling technique (SMOTE) was used to balance the classes. The authors demonstrated that their methods led to better statistical accuracy and to reduced electricity bills.


NNs are one of the most popular algorithms for peak load forecasting tasks because of their strong performance in non-linear modelling. Their flexibility is remarkable, but it is difficult to pick the right architecture and hyper-parameters for a specific problem. One of the first papers proposing a NN peak load forecasting method was produced by \cite*{dash_peak_1995}. According to the authors, NNs performed well on load forecasting problems, but they were much less performant on peak load forecasting tasks. A fuzzy NN was found to be more robust and accurate than a traditional NN structure. It involved an additional layer of fuzzification of the inputs before entering the only hidden layer of the network. 

In a more traditional set-up, \cite*{saini_artificial_2002} tested a Fully Connected Neural Network (FCNN) with different variants of back-propagation algorithms where training was conducted separately in four periods of time during a year. Their work was further developed by \cite*{saini_peak_2008}, where numerous weather variables were included (e.g., temperature, rainfall, wind speed, evaporation per day, sunshine hours and associated statistics). Similarly, different optimisation procedures were considered and it was found that an adaptive learning method based on the learning rate and momentum was the most performant. \cite*{amin-naseri_combined_2008} combined a self-organising map with a NN to find better clusters of training data to improve forecasting performance.  Some authors considered other form of networks. For instance, \cite*{abdel-aal_modeling_2006} adopted abductive networks with the aim of obtaining a better intuition and a more automated way to address peak load forecasting. In particular, these networks split the overall problem into smaller and simpler ones along the network with abductive reasoning. It is based on an automated procedure which organises the data available into different chunks and deals with them separately. 

More recently, recurrent Neural Networks (RNNs) have been used by \cite*{yu_deep_2019} in the form of Gated Recurrent Units (GRU). In particular, a dynamic time warping (DTW) analysis was used to produce the GRU inputs. The DTW distance was used to find the most similar load curve to the one observed before the targeted load curve. Assuming that subsequent load curves are also very similar, they used the subsequent load curve from the training data to encode the inputs of the GRU network. A Long Short-Term Memory (LSTM) architecture has been used by \cite*{ibrahim_lstm_2020} and was found to be more computationally efficient compared to FCNNs and other RNNs. Three statistical metrics were used to evaluate model performance: Mean Absolute Percentage Error (MAPE), Root-Mean Squared Error (RMSE) and mean bias error. In our work, statistical metrics including MAPE and RMSE will also be used to avoid introducing any bias towards a particular operational application. 

The literature on deep learning peak load forecasting is sparse, but deep learning probabilistic load forecasting is much more common (e.g., \citealp*{guo_deep_2018}, \citealp*{yang_deep_2019} and \citealp*{yang_bayesian_2020}). Such models do not explicitly focus on the DP or the IP as the objective functions used to estimate their parameters are based on demand observed at a higher frequency (intra-day). The high-frequency forecasts thus obtained can be post-processed to produce a forecast for the DP. 

Support Vector Regression (SVR) is another popular class of load forecasting method, based on structural risk minimisation instead of empirical risk minimisation as in NNs. \cite*{el-attar_forecasting_2009} used SVR in a local prediction framework. Recently, \cite*{kim_peak-load_2020} used an ensemble forecasting approach with other Machine Learning algorithms such as boosting machines, tree-based methods and bagging techniques. A compensation process based on an isolation forest is later added by analysing the predicted values of the ensemble models to detect outliers in the peak data. SVR are compared to NNs by \cite*{li_analysis_2018} for a control strategy of peak load and frequency regulation. LSTM NNs were used to forecast power load and improve the control strategy considered in this particular use case.

From this literature review, it can be concluded that a wide range of methodologies have been adopted in peak load forecasting applications. In most short-term applications, model inputs are manually chosen features that are defined at the same (daily) time resolution as the peak demand, which is the variable to be forecasted. Conversely, in long-term applications, weather variables are simulated at the original (high) resolution to produce demand forecasts at the same resolution, which are then post-processed to obtain low resolution (e.g., yearly) peak forecasts. Hence, to the best of our knowledge, the existing literature on peak forecasting has not explored methods that are able to integrate both low- and high-resolution signals in a single model. However, in the field of functional data analysis, hybrid approaches have been used for clustering and forecasting functional data (e.g., \citealp*{antoniadis_functional_2006} and \citealp*{cho_modeling_2013}). Therefore, this paper aims to exploit functional methods to tackle multi-resolution problems. From a feature engineering point of view, the goal is to automate feature extraction of high-resolution signals, that is to let the model decide which hidden features to extract from the signal. This can be done with signal processing procedures such as tensor product decomposition, wavelets or Fourier transforms \citep*{amin_feature_2015}. 

The literature review also suggests that not much effort has been directed towards forecasting the IP, which is surprising because forecasting the IP is at least as important as forecasting the DP, for the purpose of short-term smart grid management and operational planning \citep*{soman_peak_2020}. To fill this gap, the performance of multi-resolution methods will be illustrated in this paper on both a DP and an IP forecasting problem. 

\section{Multi-resolution modelling}

In this section, the multi-resolution modelling approach is introduced with its general principles. It is then developed formally and illustrated with GAMs and NNs.

\subsection{General idea}

The main idea behind multi-resolution modelling is to build a parsimonious model that is able to handle input and output variables that are available at different resolutions.
In the context of DP load forecasting, low-resolution variables (e.g., day of the week, maximum daily temperature) are observed daily, while high-resolution variables (e.g., temperatures or raw demand) are updated every hour or half-hour. Such problems are usually handled by manually placing all variables at the same resolution. In particular, one option is to take a high-resolution approach, which consists in doing the modelling at the highest available resolution, which might require interpolating some of the low-resolution variables. Such an approach often lacks in parsimony, as the low-resolution variables are brought to the higher resolution, thus increasing the size of the data that needs to be processed, while adding no extra useful information. Another option is to take a low-resolution approach, that is to transform the high-resolution variables into a set of manually chosen daily summaries or features. In this approach, the size of the data is reduced, but feature engineering is time consuming and some of the information contained in the high-resolution variables is lost in the process. 

The multi-resolution approach proposed here aims at capturing all the information contained in the high-resolution variable, while avoiding explicit feature engineering and retaining the parsimony of the low-resolution approach. To describe the multi-resolution idea more formally, let us consider $\textbf{y}_i = \{y_i(t)\}_{t\in\{1,\ldots,T\}}$ the vector of electricity demand at each time step $t > 0$ of the day $i \in \mathbb{N}$ . $T$ is the total number of daily steps (e.g., T=48 for half-hourly steps). Then, the DP of day $i$ is $\textrm{DP}_i = max(\textbf{y}_i)$ and $\textrm{IP}_i$ is the time step corresponding to $\textrm{DP}_i$. Let $\textbf{x}^{low}_i$ be the $i$-th vector of covariates observed daily and let $\textbf{x}^{high}_i$ be the corresponding vector of covariates containing information at the intra-day resolution. The multi-resolution approach exploits both sets of covariates as model inputs to obtain the forecasts of the $\hat{\textrm{DP}}_i$ or the $\hat{\textrm{IP}}_i$, that is
\begin{align}
  \hat{\textrm{DP}}_i &= \psi_1(\textbf{x}^{low}_i,\textbf{x}^{high}_i)  \\
  \hat{\textrm{IP}}_i &= \psi_2(\textbf{x}^{low}_i,\textbf{x}^{high}_i) 
\end{align}
where $\psi_1$ and $\psi_2$ represent the model for, respectively, the DP and the IP.
This general definition does not specify how the high-resolution inputs should be dealt with in practice. Several approaches could be considered, the aim being to process the information contained in a (possibly high-dimensional) signal vector, while avoiding information loss and retaining computational efficiency. In this paper, two options are considered. In particular, a description of how high-resolution covariates can be handled within GAMs and NNs is given below. 

\subsection{Particular instances of the multi-resolution approach}

The multi-resolution approach is detailed firstly for GAMs which, due to their performance and interpretability \citep*{amato_forecasting_2021}, are widely used in industry for load forecasting. Then, the multi-resolution approach is extended to NNs, which often perform well on load forecasting problems and enable the flexible handling of heterogeneous model inputs \citep*{gao_matrix_2017}. 

\subsubsection{Generalised Additive Models}

 First introduced by \cite*{hastie_generalized_1999}, GAMs are a semi-parametric extension of generalised linear models (GLMs) where the response variable, $y_i$, is assumed to follow a parametric probability distribution. That is, $y_i \sim \text{Dist}(\mu_i, \bm \theta)$ where $\mu_i$ and $\bm \theta$ are model parameters. While the elements of $\bm \theta$ do not depend on $i$, parameter $\mu_i$ is modelled as follows \citep*{wood_generalized_2017}:
\begin{equation}
    g(\mu_{i})=\mathbf{x}_{i}^T \bm{\gamma}+\sum_{j} f_{j}(\mathbf{x}_{i}),
\end{equation}
where $g$ is a monotonic transformation, which is simply the identity function in this paper. Two separate terms can be distinguished on the right-hand side of this equation: a parametric part $\mathbf{x}_{i}^T \bm{\gamma}$, where $\mathbf{x}_{i}$ is a vector of covariates while $\bm \gamma$ is a vector of regression coefficients, and a non-parametric part $\sum_{j} f_{j}(\bm{x}_{i})$ which is a sum of smooth functions of covariates. The smooth effects are built via linear combinations of $K_j$ basis functions, while the corresponding basis coefficients are penalised via generalised ridge penalties. The strength of the penalties is controlled via smoothing hyper-parameters, which are selected using criteria such a generalised cross-validation.

In the context of forecasting $\textrm{DP}_i$, it is interesting to consider for $\text{Dist}(\mu_i, \bm \theta)$ a generalised extreme value (GEV) distribution. In fact, the GEV model is asymptotically justified for block-maxima as $T \rightarrow \infty$ \citep*{jacob_forecasting_2020}. Thus, when enough steps are available throughout the day, the GEV distribution is particularly attractive for modelling the DP. The scaled-T (a scaled version of Student's t) distribution provides an alternative, which is particularly suited for heavy tailed data such as peak load. The Gaussian distribution can be used as a baseline model. As for the IP, an ordered categorical (ocat) distribution based on a logistic regression latent variable is used. All of these distributions as well as GAM building and fitting methods are implemented in the \textit{mgcv} R package \citep*{wood_mgcv_2020}.

Within the additive structure of GAMs, $\mathbf{x}^{low}_i$ and $\mathbf{x}^{high}_i$ can be treated as inputs for different smooth functions. The elements of $\mathbf{x}^{low}_i$ can be handled via separate standard smooth effects, which take scalars as inputs, while the joint effect of several elements of $\mathbf{x}^{low}_i$ can be captured via standard multivariate smooth effects. However, the $\mathbf{x}^{high}_i$ covariates have to be treated via functional smooth effects. The latter are smooth functions which take the vectors of high-resolution covariates as inputs and output a scalar. Therefore, functional GAMs permit the handling of each covariate at its original resolution, thus avoiding interpolation and guaranteeing parsimony.

In addition to the principle of parsimony, the goal is also to retain the time dependence of the covariates. In fact, it is important to ensure that the model is aware that each element of the high-resolution covariates has a different impact on the peak load distribution, as it belongs to a different time of day. A way to retain the time dependence of each high-resolution series of covariates is to make them interact with the time of day sequence via tensor product effects. Such effects can easily be integrated in GAMs, as explained in the following.

In continuous time, the smooth effect for a high-resolution (functional) covariate, $x_i(u)$, can be written as follows:
\begin{align}
f(x_i) = \int_{0}^{T}\phi({x}_i(u),u)du
\end{align}
where $\phi$ is the time-dependent effect of the covariate, which needs to be estimated, while $u$ is the time of day. In practice, on the $i$-th day, ${x}_i(u)$ is observed at $F$ discrete instants $0 \leq t_1 \leq \cdots \leq t_F \leq T$ and the corresponding values of ${x}_i(u)$ are stored in the vector $\bm x_i$. Hence, approximating the integral with a summation and constructing $\phi$ via a tensor product expansion leads to:
\begin{align}
\hat{f}(\bm{x}_i) & = \sum_{r=1}^{F}\hat{\phi}({x}_i(t_r),t_r) \nonumber\\
                      & = \sum_{r=1}^{F} \sum_{k=1}^{K} \sum_{l=1}^{L} \beta_{kl}a_k({x}_i(t_r))b_{l}(t_r)
\end{align}
where $\{a_{k}\}_{(k)\in\{1,\ldots,K\}}$ and $\{b_{l}\}_{(l)\in\{1,\ldots,L\}}$ are known spline basis functions and $\{\beta_{kl}\}_{(k,l) \in\{1,\ldots,K\} \times \{1,\ldots,L\}}$ are parameters to be estimated. By using such effects, high-resolution information can be parsimoniously incorporated into the model, while retaining the temporal information contained in the covariates. 

\subsubsection{Neural Networks}

NNs are convenient machine learning algorithms to implement a multi-resolution model. In fact, common architectures such as Convolutional Neural Networks (CNN) and RNNs already make use of inputs from different scales. Recent work was undertaken to make tensor inputs available for multi-layer perceptrons with MatNet \citep*{gao_matrix_2017} which further shows their versatility. From scalars to tensors, the flexibility of NNs is hard for other machine learning models to compete with.

A FCNN or CNN architecture, without its output layer, can be generally written as follows:
\begin{align}
   H_k(\mathbf{x},\Theta) = h_k (\ldots h_3(h_2(h_1(\mathbf{x},\theta_1),\theta_2),\theta_3)\ldots, \theta_k ) 
\end{align}
where k is the number of hidden layers of the NN, ${h_i}_{,i \in \{1 \ldots k\}}$ are the transformations made by the hidden layers (e.g., linear operation, activation and dropout) and $\Theta = \{\theta_i\}_{,i \in \{1 \ldots k\}}$ is the sequence of parameter vectors (weights and biases). In a multi-resolution approach, one part of the architecture will contain low-resolution information feeding a FCNN branch and the other one will contain the reshaped high-resolution data feeding a CNN or RNN branch. In this paper, only CNNs were considered in depth for this latter branch, with the lags of the response provided as model inputs. The CNN enables a very close replication of the tensor product construction used for GAMs, thus creating a consistent set-up for comparing both algorithms.

\begin{figure}[H]
    \centering
    \includegraphics[width=\linewidth,keepaspectratio]{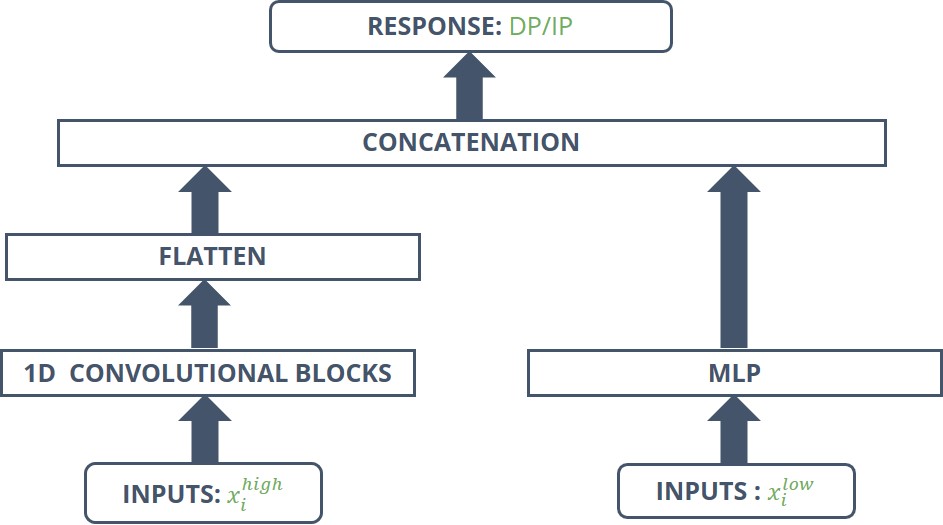}
    \caption{Multi-resolution architecture for NNs with a Multi-Layer Perceptron (MLP) taking low-resolution inputs, and a CNN with high-resolution inputs}
    \label{MRNN}
\end{figure}

Even though the CNN and FCNN branches do not have similarly shaped inputs and outputs, the unit shapes can be transformed along the network to interact and be brought together without losing consistency. This process consists in flattening the tensor shapes in order to bounce back on vectorial inputs within some layer of the network. It is precisely this flexibility that can be leveraged to build a multi-resolution architecture (Figure \ref{MRNN}).
More precisely, the CNN branch contains one convolutional block for each of the high-resolution time series. In this way, each tensor product of the GAM formula can find its equivalent in the CNN branch of the network. In fact, the multi-resolution NN architecture can be concisely written as follows:  
\begin{align}
    \mathbf{\mu}_i = F_j(H_k(\mathbf{x}_{low},\Theta), H'_l(\mathbf{x}_{high},\Theta'))
\end{align}
In (7), $H_k$ is the FCNN which handles low-resolution terms while $H'_l$ is the CNN which deals with the high-resolution information. Then, in the final part of the network, both outputs are concatenated (after flattening the CNN branch) and enter another FCNN $F_j$ which can be reduced to the output layer when $j = 1$. Here, $\mu_i$ is the mean of the random output variable considered. This multi-resolution architecture is summarised in Figure \ref{MRNN}.

\section{Experiments}

On the DP and the IP forecasting tasks, the multi-resolution approach is compared to two alternative modelling approaches: a high-resolution approach and a low-resolution approach (Figure \ref{High-, low- and multi-resolution modelling setting}). The low-resolution approach uses inputs aggregated at the daily level (e.g., maximum daily temperature, day of the week) to forecast the DP and the IP separately. The high-resolution approach uses inputs at the half-hourly level to forecast the half-hourly demand and then it extracts the DP and the IP by taking the maximum of the half-hourly forecasted values and the corresponding time of day. Therefore, the high-resolution approach leverages all the information available by taking half-hourly inputs and outputs while the low-resolution approach directly models the variables of interest (DP and IP) with less parameters to be estimated. The multi-resolution approach can be seen as a compromise, aimed at integrating the advantages of both approaches, and the following experiments are designed to assess whether it can outperform them. 

\begin{figure}[H]
    \centering
    \includegraphics[width=\linewidth,keepaspectratio]{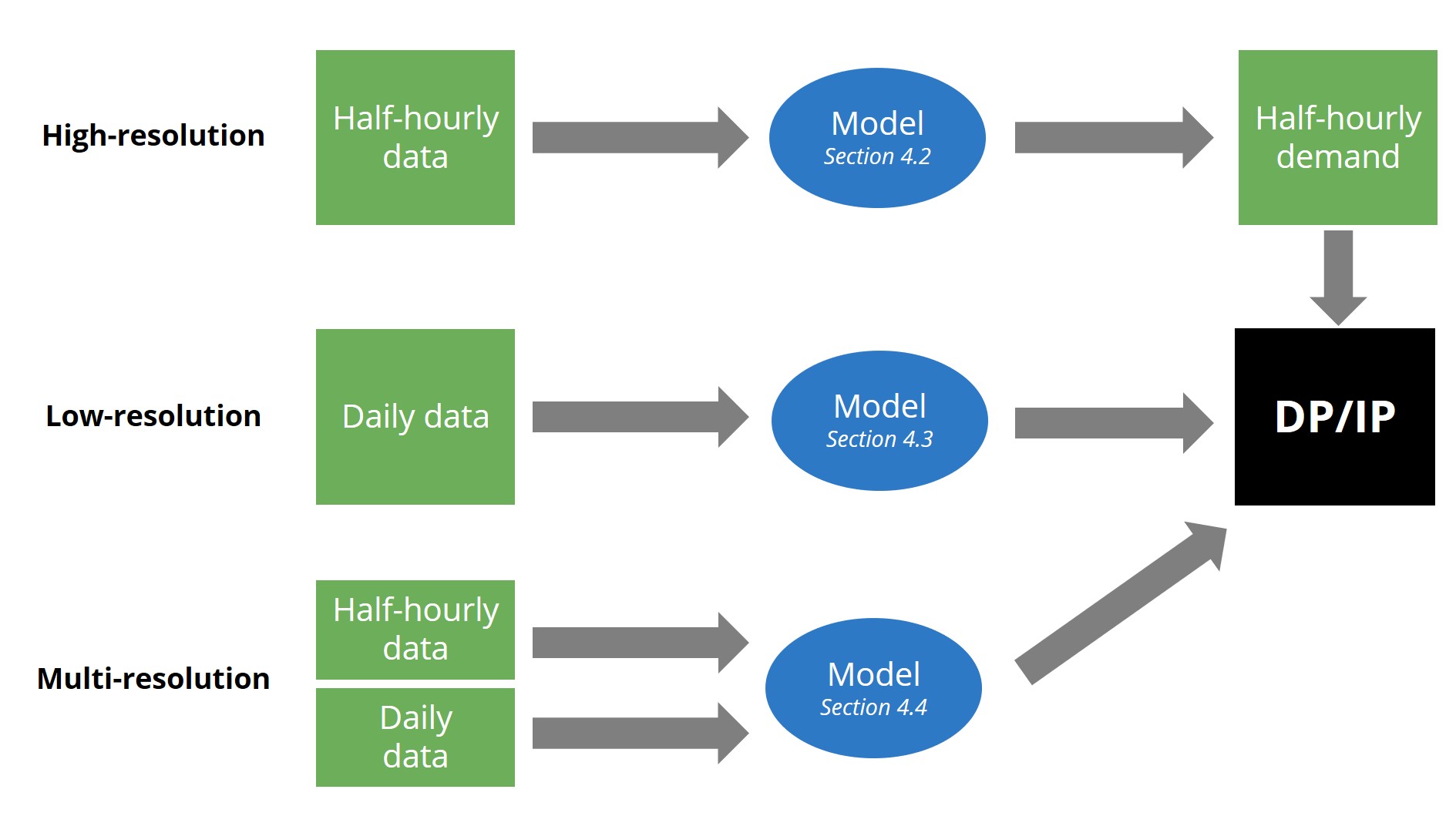}
    \caption{The different modelling settings compared in this work}
    \label{High-, low- and multi-resolution modelling setting}
\end{figure}

The comparison includes baseline models: a naive persistence model, which simply consists of forecasting the DP and the IP based on the value taken by the target variable on the previous day; a low-resolution ARIMA (on daily peaks with horizon 1); a high-resolution ARIMA aggregated forecast composed of 48 ARIMA models, each fitted on the half-hourly load of a specific time of day with horizon 1. That is, the high-resolution ARIMA produces 48 forecasts at horizon 1 instead of one forecast at horizon 48. All ARIMA models are fitted using the \cite*{hyndman_automatic_2007} algorithm without using exogenous information.

The performance metrics chosen for DP models are the mean absolute percentage error (MAPE), the mean absolute error (MAE) and the root mean squared error (RMSE). As for IP models, the same metrics are used except for the MAPE, which is substituted with a relaxed accuracy (R-Accuracy) metric in the form of a binary loss function (equal to 1 if the IP forecasted is more than 2 instants away from the observed IP and 0 if it is within 2 instants of the observed IP). While the R-Accuracy metric is also relevant in operational settings where it is crucial to know the IP within a small time window, the RMSE and the MAE penalise forecasts proportionally to their distance from the observed IP.

\begin{figure}[H]
    \centering
    \includegraphics[width=\linewidth,keepaspectratio]{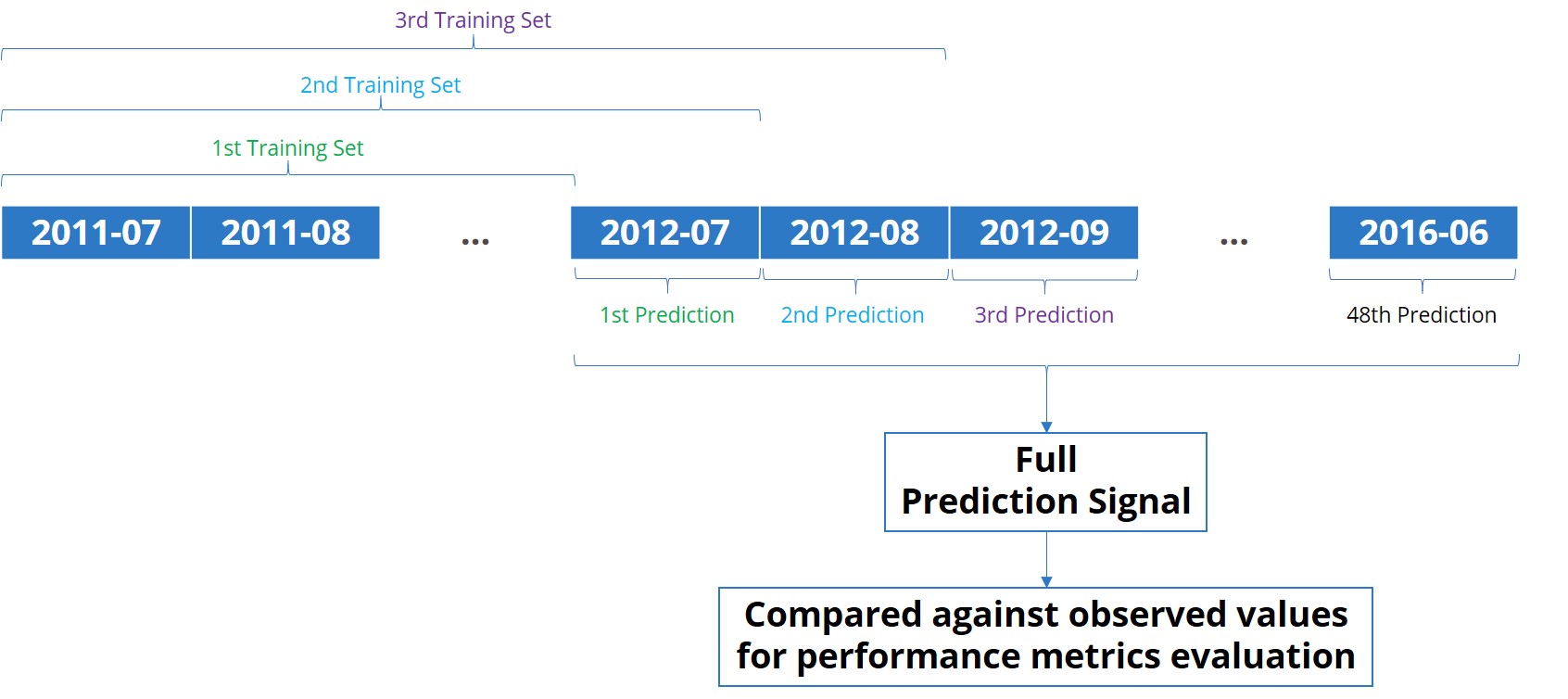}
    \caption{Rolling-origin forecasting procedure}
    \label{Rolling Origin}
\end{figure}

A rolling-origin forecasting procedure is used to replicate a realistic short-term load forecasting set-up (Figure \ref{Rolling Origin}). Model parameters are updated on a monthly basis with consolidated data since, in an operational setting, threats to data validity and computational constraints can emerge when refitting a model too often using real-time data.


The data used in the experiments is the half-hourly load consumption (total national demand) between 2011-07-01 00:00:00 and 2016-06-30 23:30:00, available via the UK \cite*{nationalgrid_eso_2021} website. Temperature data at different locations (London, Sheffield, Manchester, Leeds, Cardiff, Bristol, Birmingham, Liverpool, Crosby and Glasgow) was downloaded from the \cite*{noaa_national_2021} website. The temperature data is at an hourly resolution. It is interpolated (natural cubic spline interpolation) to obtain half-hourly data. Furthermore, demographic information $pop_s$ is compiled around each station $s$ and a weighted mean temperature is calculated as follows:
\begin{equation*}
    \mathrm{temp}(t) = \frac{1}{\sum_{a=1}^{10} pop_s} \sum_{s=1}^{10} pop_s T_{s,t}
\end{equation*}
where $T_{s,t}$ is the temperature recorded at time $t$ by station $s$ and $\mathrm{temp}(t)$ is the weighted mean temperature which will be used in the modelling experiments. An exponentially smoothed version of the weighted mean will also be included in the model features. It was computed using a smoothing parameter equal to 0.95.

\subsection{High-resolution approach}

Forecasting the electricity hourly or half-hourly demand is a problem that has been extensively studied in the literature \citep*{kuster_electrical_2017}. It is well known that a common driver of electrical load is weather and in particular temperature. In addition, calendar information can be used to explain the seasonal variation of the demand. Finally, lagged demand values are highly informative for the subsequent values. These variables are summarised in Table 1.

\begin{table}[H]
\caption{High-resolution model inputs}
\centering
\begin{tabular}{@{}cccc@{}}
\toprule
Type                      & Name   & Unit        & Description                                                \\ \midrule
\multirow{2}{*}{Weather}  & temp   & {[}C°{]}    & Half-hourly temperature                                    \\ \cmidrule(l){2-4} 
                          & temp95 & {[}C°{]}    & Half-hourly smoothed temperature                           \\ \midrule
\multirow{3}{*}{Calendar} & dow    & Categorical & Day of the week                                            \\ \cmidrule(l){2-4} 
                          & toy    & None        & Time of year (between 0 and 1)                             \\ \cmidrule(l){2-4} 
                          & t    & Categorical & Time of day (between 0 and 47)                             \\ \midrule
Lag                       & load24 & {[}$10^{1}$ GW{]}    & Half-hourly load on the previous day \\ \midrule
Output                    & load   & {[}$10^{1}$ GW{]}    & Half-hourly load                                           \\ \bottomrule
\end{tabular}%
\end{table}

The GAM chosen to implement this approach is $y_i(t) \sim N(\mu_i(t), \sigma^2)$ where the mean of the Gaussian distribution is modelled by:
\begin{align}
   \mu_{i}(t) = & \, \psi_{1}(\mathrm{dow}_{i})+ \psi_{2}(\mathrm{t}) + f_{1}^{20}(\mathrm{toy}_{i}(t)) + f_{2}^{20}(\mathrm{temp}_{i}(t)) + f_{3}^{24}(\mathrm{temp95}_{i}(t))  \nonumber \\ 
               & +\mathrm{ti}_{1}^{5,5}(\mathrm{temp}_{i}, t)+\mathrm{ti}_{2}^{5,5}(\mathrm{temp95}_{i}(t), t)+\mathrm{ti}_{3}^{5,5}(\mathrm{load24}_{i}(t), t) \\ \nonumber
               & + \mathrm{ti}_{4}^{5,5}(\mathrm{toy}_{i}(t), t)
\end{align}
In (8), the $\psi$ functions are parametric effects, while the $f$ functions are univariate smooth effects and the $\mathrm{ti}$ functions are bivariate tensor product smooth interactions. The number of basis functions used is indicated in the exponents. For instance, $f_{1}^{20}$ uses 20 basis functions and $\mathrm{ti}_{1}^{5,5}$ uses 5 basis functions for each marginal. Thin-plate spline bases are used to build all smooth effects \citep*{wood_thin_2003}. The model structure (8) was decided on the basis of previous experience in the field and the statistical significance of each effect. 

There are many NN architectures which could be considered for this problem. We want an architecture with the minimum number of layers possible and using the same model inputs as the GAM. Adding too many layers would lead to a drastic difference in degrees of freedom between the NN and the GAM which is not realistic in a short-term load forecasting scenario. Furthermore, as we are not in the big data regime, adding too many layers may actually worsen the performance of the network.

Given that the universal approximation theorem (\citealp*{cybenko_approximation_1989} and \citealp*{hornik_approximation_1991}) guarantees that a two-layer FCNN can approximate any measurable function on a compact support, a FCNN carefully built can approximate any non-linear function of the input variables with only one hidden layer. Therefore, a FCNN architecture was used to build an NN analogue of the high-resolution GAM baseline model.

\begin{figure}[H]
    \centering
    \includegraphics[width=\linewidth,keepaspectratio]{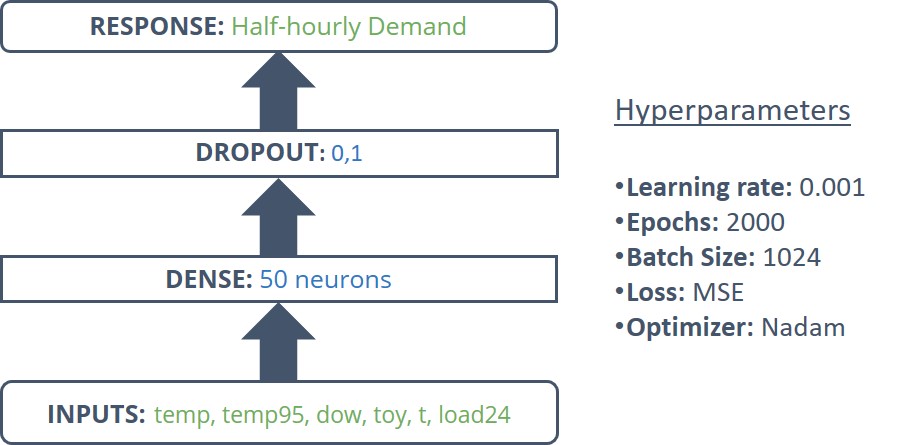}
    \caption{High-resolution FCNN architecture (input variable names are detailed in Table 1)}
    \label{HRFCNN}
\end{figure}

In practice, there is no bound for the number of hidden units, which can lead to poor generalisation of the model when assessed on the test set. Therefore, a dropout layer was added after the hidden layer to foster the network generalisation. The outcome of the optimisation of hyperparameters led to the architecture shown in Figure \ref{HRFCNN}, which contains 50 neurons in the hidden layer and a dropout layer with a 10\% dropout rate.

After obtaining the half-hourly demand forecast for the GAM and the NN, $\hat{\textrm{DP}}_i$ is estimated as the maximum daily value forecasted and $\hat{\textrm{IP}}_i$ is estimated as the half-hour of the day during which $\hat{\textrm{DP}}_i$ occurred.

\subsection{Low-resolution approach}

\begin{table}[H]
\caption{Low-resolution model inputs}
\resizebox{\textwidth}{!}{%
\begin{tabular}{@{}cccc@{}}
\toprule
Type                      & Name      & Unit        & Description                        \\ \midrule
\multirow{4}{*}{Weather}  & tempMax   & {[}C°{]}    & Daily maximum temperature          \\ \cmidrule(l){2-4} 
                          & temp95Max & {[}C°{]}    & Daily maximum smoothed temperature \\ \cmidrule(l){2-4} 
                          & tempMin   & {[}C°{]}    & Daily minimum temperature          \\ \cmidrule(l){2-4} 
                          & temp95Min & {[}C°{]}    & Daily minimum smoothed temperature \\ \midrule
\multirow{2}{*}{Calendar} & dow       & Categorical & Day of the week                    \\ \cmidrule(l){2-4} 
                          & toy       & None        & Time of year (between 0 and 1)     \\ \midrule
\multirow{2}{*}{Lag}      & DP24      & {[}$10^{1}$ GW{]}   & Previous day peak demand           \\ \cmidrule(l){2-4} 
                          & IP24      & Categorical & Previous day instant of peak       \\ \midrule
Output                    & DP or IP        & {[}$10^{1}$ GW{]} or Categorical   & Daily demand peak or Daily instant of peak                  \\ \bottomrule
\end{tabular}%
}
\end{table}

In the low-resolution approach, all input variables are at the daily resolution (Table 2). Here several distributions could be considered for GAMs. In particular, the scaled-T distribution, which is particularly suited for heavy tailed data, as well as the GEV family, which encompasses several extreme value distributions (Weibull, Gumbell and Fréchet), are used to model the DP. For the IP forecasting task, the ordered-logit model implemented in the \textit{mgcv} R package \citep*{wood_mgcv_2020} is used. The low-resolution GAM can be written as follows:
%
\begin{align}
   \mu_{i} =& \psi_{1}(\mathrm{dow}_{i})+ f_{1}^{10}(\mathrm{IP24}_{i}) + f_{2}^{20}(\mathrm{toy}_{i}(t)) +   f_{3}^{20}(\mathrm{DP24}_{i})   \nonumber \\ 
               & + f_{4}^{20}(\mathrm{tempMax}_{i}(t)) + f_{5}^{20}(\mathrm{temp95Max}_{i}(t))  \\ \nonumber
               & + f_{6}^{20}(\mathrm{tempMin}_{i}(t)) + f_{7}^{20}(\mathrm{temp95Min}_{i}(t))
\end{align}
%
For the DP, $\mu_{i}(t)$ is the location parameter of the distributions estimated, the other parameters are assumed to be constants. For the IP, $\mu_{i}(t)$ is also the location parameter of a latent logistic distribution. Cut-off points are estimated in the course of model fitting and do not depend on the covariates. See \cite*{wood_smoothing_2016} for details. 

\begin{figure}[H]
    \centering
    \includegraphics[width=\linewidth,keepaspectratio]{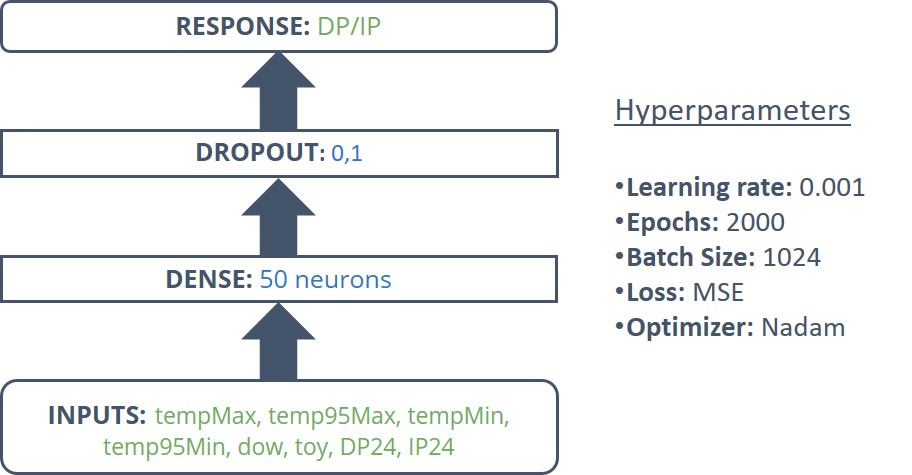}
    \caption{Low-resolution FCNN architecture (input variable names are detailed in Table 2)}
    \label{LRFCNN}
\end{figure}

The same FCNN architecture as for the high-resolution approach was used (Figure \ref{LRFCNN}). The only difference between them is the response variable which here is directly the DP or the IP. Furthermore, the hyperparameters chosen are different. In particular the number of epochs and the batch size are much larger. The response structure for the DP is 1 neuron with a ReLU activation while 48 neurons are used for the IP. Instead of the traditional softmax output used in classification problems, an ordinal output structure, more suited to model the IP, is implemented as formalised by \cite*{jianlin_cheng_neural_2008}. The observed response is structured as a vector of 1 and 0. If the peak was observed at $t \in \{1,\ldots,T\}$ all neurons before and including the t-th one will be 1 and all neurons after will be 0. Therefore, sigmoidal activation functions are used.

\subsection{Multi-resolution approach}

The multi-resolution GAMs leverage the same level of information for model inputs as in the high-resolution GAMs. In addition, the directly targets the DP response variable as in the low-resolution approach.
\begin{table}[H]
\caption{Multi-resolution model inputs}
\resizebox{\textwidth}{!}{%
\begin{tabular}{@{}cccc@{}}
\toprule
Type                      & Name     & Unit        & Description                                  \\ \midrule
\multirow{2}{*}{Weather}  & matTem   & {[}C°{]}    & Vector of half-hourly temperatures           \\ \cmidrule(l){2-4} 
                          & matTem95 & {[}C°{]}    & Vector of half-hourly smoothed temperatures  \\ \midrule
\multirow{3}{*}{Calendar} & dow      & Categorical & Day of the week                              \\ \cmidrule(l){2-4} 
                          & toy      & None        & Time of year (between 0 and 1)               \\ \cmidrule(l){2-4} 
                          & matInt   & Categorical & Vector of time steps (between 0 and 47)      \\ \midrule
Lag                       & matLag   & {[}$10^{1}$ GW{]}    & Vector of half-hourly load from previous day \\ \midrule
Output                    & DP or IP        & {[}$10^{1}$ GW{]} or Categorical   & Daily demand peak or Daily instant of peak                               \\ \bottomrule
\end{tabular}%
}
\end{table}
Tensor products defined in Section 3.2.1 are used to capture high-resolution information. The \textit{mat} covariates presented in Table 3 are matrices of dimension $(N \times 48)$, $N$ being the number of observations of the response variable DP. The multi-resolution GAM model is:
\begin{align}
   \mu_{i} = & \, \psi_{1}(\mathrm{dow}_{i})+ f_{1}^{20}(\mathrm{toy}_{i}) + ti_{1}^{15,10}(\mathrm{matTem}_{i}, \mathrm{matInt}_{i}) \nonumber \\  
   & + ti_{2}^{5,5}(\mathrm{matTem95}_{i}, \mathrm{matInt}_{i}) + ti_{3}^{5,5}(\mathrm{matLag}_{i}, \mathrm{matInt}_{i})
\end{align}
Unlike previous approaches, IP and DP lags are not directly included as they can be captured by the model through the $ti_{3}$ tensor interaction. As for the low-resolution approach, Gaussian, scaled-T and GEV distributions are considered for the DP and the ordered categorical distribution for the IP. 

\begin{figure}[H]
    \centering
    \includegraphics[width=\linewidth,keepaspectratio]{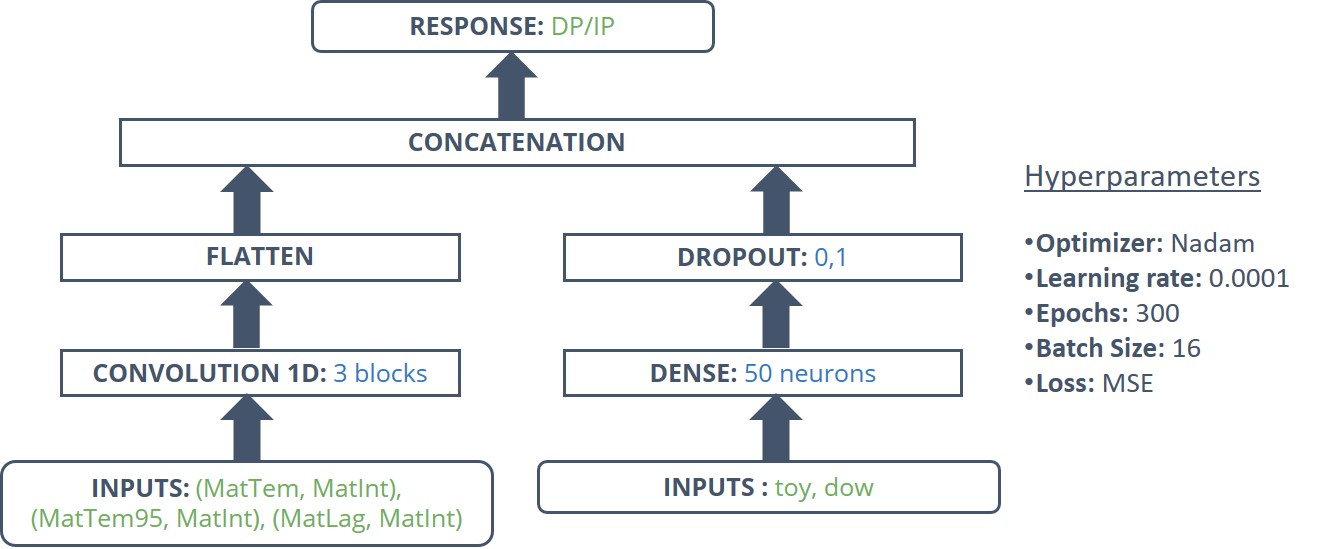}
    \caption{Multi-resolution CNN architecture (input variable names are detailed in Table 3)}
    \label{MRCNN}
\end{figure}

For the multi-resolution NN, the tensor product interactions will be replaced by convolution layers. The mechanism looked for through these convolution layers is essentially the same as for tensor products: extracting high-resolution information to directly model the DP or the IP. The high-resolution (half-hourly) data will be passed on to the convolution layers while the low-resolution (daily) data will go through the same FCNN architecture used in the previous approaches. As shown in Figure \ref{MRCNN}, these two sections of the architecture are then concatenated to produce the final forecast of the DP load. The output structure for the DP and the IP are the same as detailed in Section 4.3 with one neuron for the DP and 48 neurons for the IP. 

The convolutions used for the high-resolution information are 1D convolutions on two channels. Usually, only one convolution funnel is used to capture interactions between all inputs. Here, each tensor product interaction will be replicated as a unique convolutional block. Thus, three convolution blocks will independently extract the three high-resolution terms: matTem, matTem95 and matLag. The second channel of each block is the matrix containing the vectors of time steps matInt.

\section{Results}

The performance of the models for the DP and the IP forecasting tasks is evaluated using three statistical metrics. As a rolling-origin forecasting procedure was chosen, a transitional regime can be observed in the first few iterations, particularly for NNs, which usually perform better with a large amount of training data. Therefore, Table 4 (DP) and Table 5 (IP) present the models' performances on the last year of data, that is, from 2015-07-01 to 2016-06-30 included.

\begin{table}[H]
\caption{Performance on the last year of data for the DP (best model and associated metrics are in \textbf{bold})}
\centering
\begin{tabular}{@{}ccccc@{}}
\toprule
\multirow{2}{*}{Resolution}       & \multirow{2}{*}{Model} & \multicolumn{3}{c}{Metrics} \\ \cmidrule(l){3-5} 
                                  &                        & MAPE [\%]     & MAE [MW] & RMSE [MW]        \\ \midrule
NA                                & Persistence            & 4.38         & 23.0   & 34.3         \\ \midrule
\multirow{4}{*}{High}  
                                  & ARIMA           & 4.08 & 21.0  & 27.8        \\ \cmidrule(l){2-5} 
                                  & Gaussian GAM          & 2.43   & 13.0  & 15.5          \\ \cmidrule(l){2-5} 
                                  & FCNN                   & 1.47        & 7.77  & 10.3           \\ \midrule
\multirow{7}{*}{Low}   & ARIMA           & 3.85      & 20.0  & 26.7 \\ \cmidrule(l){2-5} 
                                  & Scat GAM               & 1.92          & 10.5 & 12.9          \\ \cmidrule(l){2-5} 
                                  & GEV GAM             & 2.67        & 14.5  & 16.9           \\ \cmidrule(l){2-5} 
                                  & Gaussian GAM           & 2.26       & 12.3  & 14.4           \\ \cmidrule(l){2-5} 
                                  & FCNN                   & 2.11          & 11.2  & 14.4          \\ \midrule
\multirow{5}{*}{Multi} & GEV GAM             & 1.52         & 8.19  & 10.3           \\ \cmidrule(l){2-5} 
                                  & \textbf{Scat GAM}      & \textbf{1.41} & \textbf{7.55}  & \textbf{9.59}   \\ \cmidrule(l){2-5} 
                                  & Gaussian GAM           & 1.42         & 7.65  & 9.63         \\ \cmidrule(l){2-5} 
                                  & CNN                    & 1.56         & 8.44 & 10.5        \\ \bottomrule
\end{tabular}%
\end{table}

With the exception of the high-resolution FCNN, the multi-resolution models perform better than the alternatives across all metrics (Table 4). The relative strong performance of the high-resolution FCNN can be explained by the large amount of high-resolution data available, which suits the needs of NNs. Further, the FCNN contains more parameters to estimate and is thus more flexible than the high-resolution GAMs, which require the user to manually specify how the effect of each input variable should be modelled. Nevertheless, the best model on all metrics is the scaled-T GAM, built using the multi-resolution approach. The GEV GAM performed worse than the other distributions, which is surprising given that the GEV distribution is asymptotically justified for BM. Interestingly, the shape parameter estimated was found to be close to 0, under which value the GEV model is simply a Gumbel distribution. 

\begin{table}[H]
\caption{Performance on last year of data for the IP (best model and associated metrics are in \textbf{bold})}
\resizebox{\textwidth}{!}{%
\centering
\begin{tabular}{@{}cccccc@{}}
\toprule
\multirow{2}{*}{Resolution}                        & \multirow{2}{*}{Model}                  & \multicolumn{3}{c}{Metrics}                     \\ \cmidrule(l){3-5} 
                                  &                        & R-Accuracy [\%]       & MAE [half-hour]    & RMSE [half-hour]        \\ \midrule
NA                     & Persistence        & 79.4              &  2.49      & 5.36  \\ \midrule
\multirow{2}{*}{High}  & Gaussian GAM       & 82.6      & 2.01       & 4.59 \\ \cmidrule(l){2-5} 
                       & FCNN               & 81.8      & 1.93      & 4.39 \\ \midrule
\multirow{2}{*}{Low}   & Ocat GAM           & 79.1      &  2.11      & 4.22  \\ \cmidrule(l){2-5} 
                       & FCNN               & 83.2      &  1.94      & 4.40 \\ \midrule
\multirow{2}{*}{Multi} & Ocat GAM           & 79.4      &  2.01 & 4.08 \\ \cmidrule(l){2-5} 
                       & \textbf{CNN}       & \textbf{83.5}    & \textbf{1.70}  & \textbf{3.85} \\ \bottomrule
\end{tabular}%
}
\end{table}


IP multi-resolution models have a similar or better performance than high- and low-resolution alternatives within the same model class on the MAE and RMSE metrics (Table 5) and the multi-resolution CNN is the best model under all metrics.  However, the metrics are affected by high sampling variability. The reasons for this are detailed later in this section, where we also argue that the mediocre performance of ocat GAMs for IP forecasting is not fundamental, but attributable to the insufficient flexibility of the specific ocat parametrisation adopted here. 

\begin{figure}[H]
    \centering
    \begin{tabular}{c|c}
    \textbf{DP} & \textbf{IP} \\
        \includegraphics[width=0.45\linewidth,keepaspectratio,page=1]{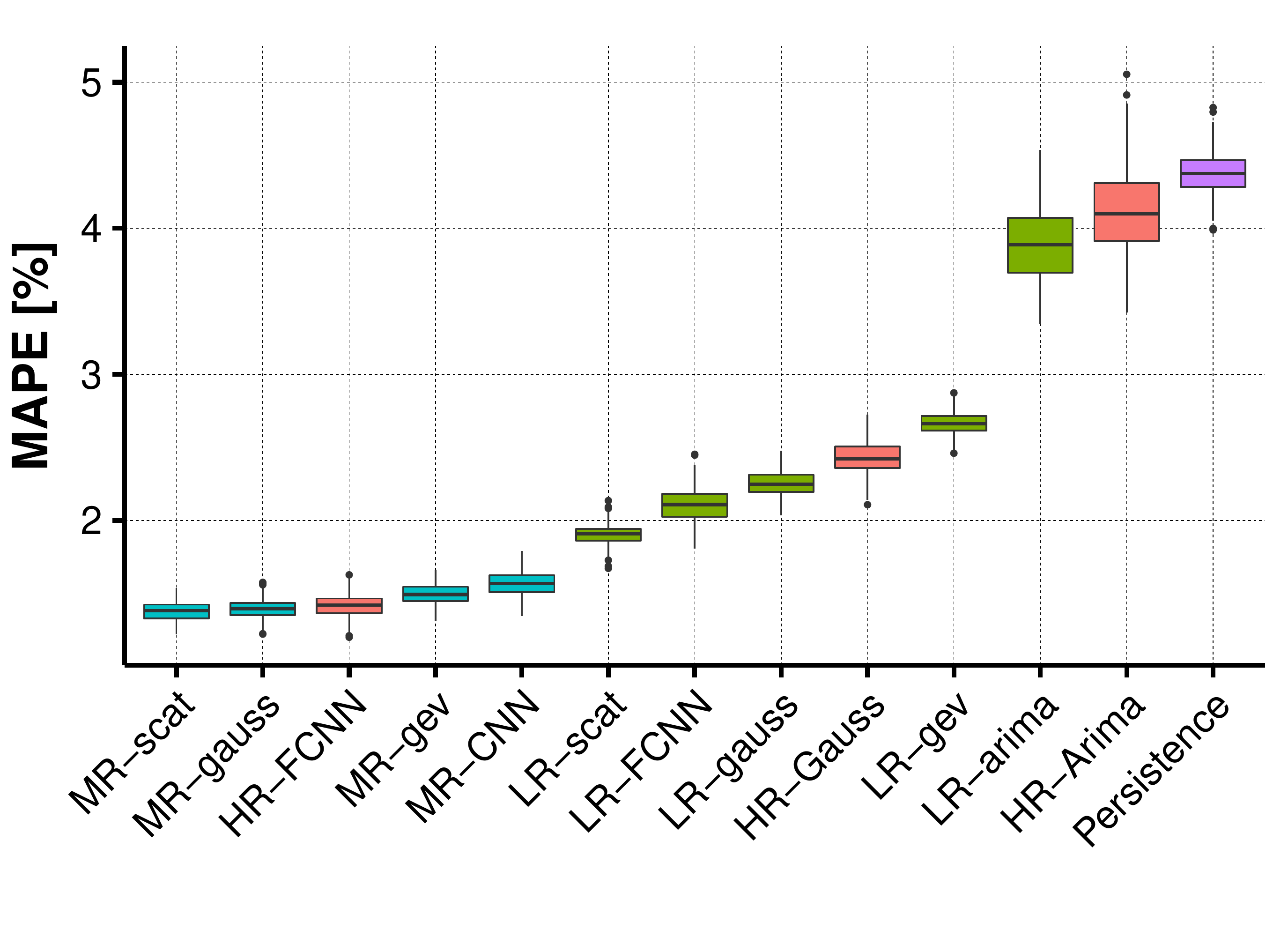} & \includegraphics[width=0.45\linewidth,keepaspectratio,page=2]{IP-block-bootstrap.pdf} \\
        (a) & (d)  \\[6pt]
        \includegraphics[width=0.45\linewidth,keepaspectratio,page=2]{DP-block-bootstrap.pdf} &
        \includegraphics[width=0.45\linewidth,keepaspectratio,page=3]{IP-block-bootstrap.pdf} \\
         (b) & (e)   \\[6pt]
        \includegraphics[width=0.45\linewidth,keepaspectratio,page=3]{DP-block-bootstrap.pdf} &
        \includegraphics[width=0.45\linewidth,keepaspectratio,page=4]{IP-block-bootstrap.pdf} \\
         (c) & (f)   \\[6pt]
         \multicolumn{2}{c}{\includegraphics[width=0.65\linewidth,keepaspectratio,page=7]{legend-block-bootstrap.pdf}}
     \end{tabular}
\caption{Block-bootstrap boxplots of the three metrics considered for the DP models (a), (b), (c) and IP models (d), (e), (f) on the last year of data}
\label{boxplots}
\end{figure}

To quantify the variability of the performance metrics considered so far, we used block-bootstrap resampling. As described by \cite*{forecast_eval}, for a test set of size $N$, we sample with replacement data blocks of fixed size $B=7$ (i.e., one week) to obtain an  evaluation sets of size $N$. Repeating this procedure $K$ times creates $K$ metric samples, which can be used to estimate the metric's sampling variability. In particular, Figure \ref{boxplots} shows block-bootstrapped boxplots for all metrics and models on the last year of data. Figures \ref{boxplots} (a-c) clearly demonstrate that the improvement obtained by adopting a multi-resolution approach is substantial and robust within the GAM model class. The HR-FCNN is competitive in terms of prediction but, as we discuss below, it is not easily interpretable and does not have the computational advantages of multi-resolution GAMs. For the IP problem, Figures \ref{boxplots} (e-d) make clear that the sampling variability is substantial (reasons for this are discussed below).

\begin{figure}[H]
    \centering
    \begin{tabular}{>{\centering\arraybackslash}m{0.45\linewidth} >{\centering\arraybackslash}m{0.45\linewidth} }
        \includegraphics[width=\linewidth,keepaspectratio]{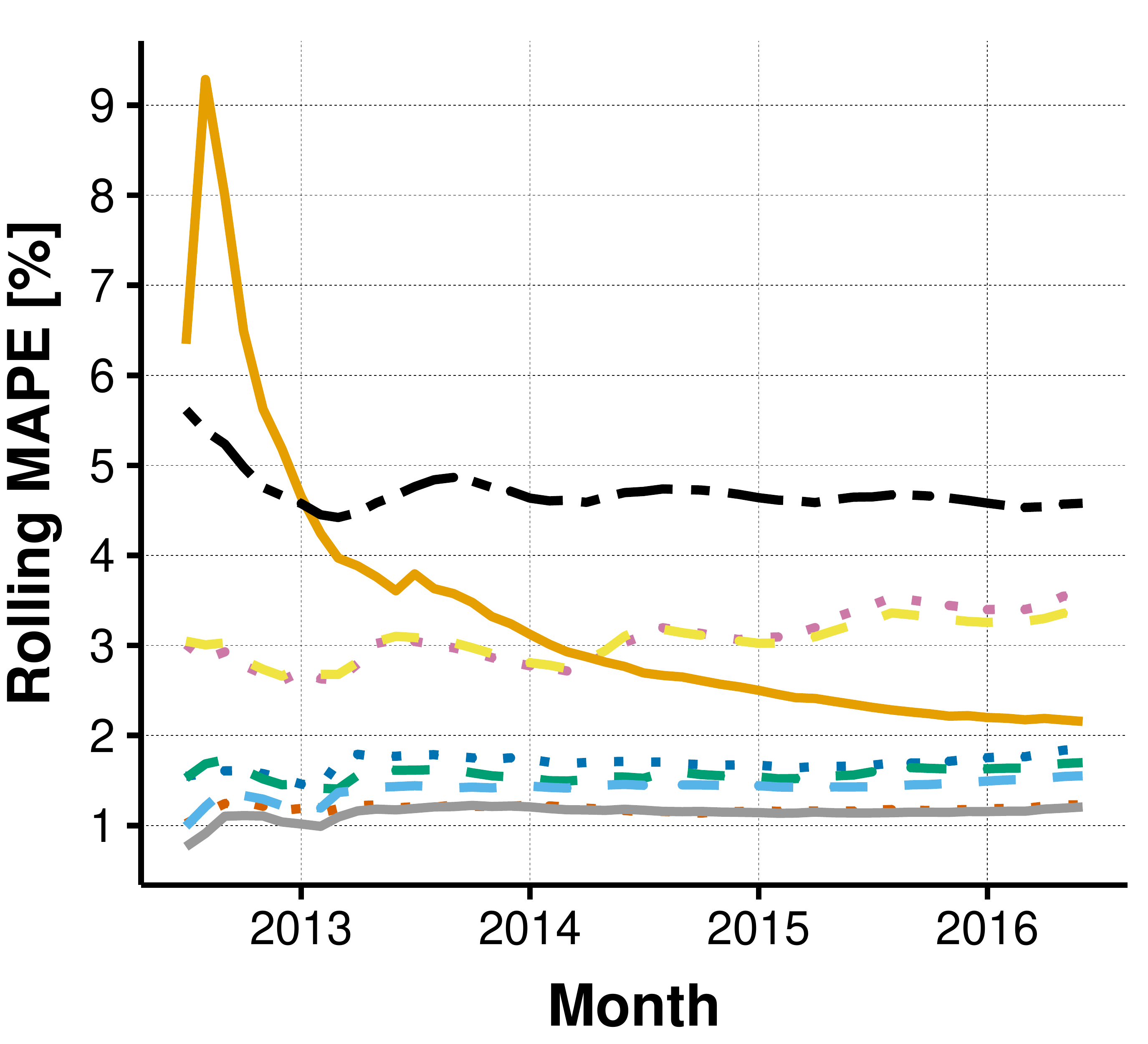} &
        \includegraphics[width=\linewidth,keepaspectratio]{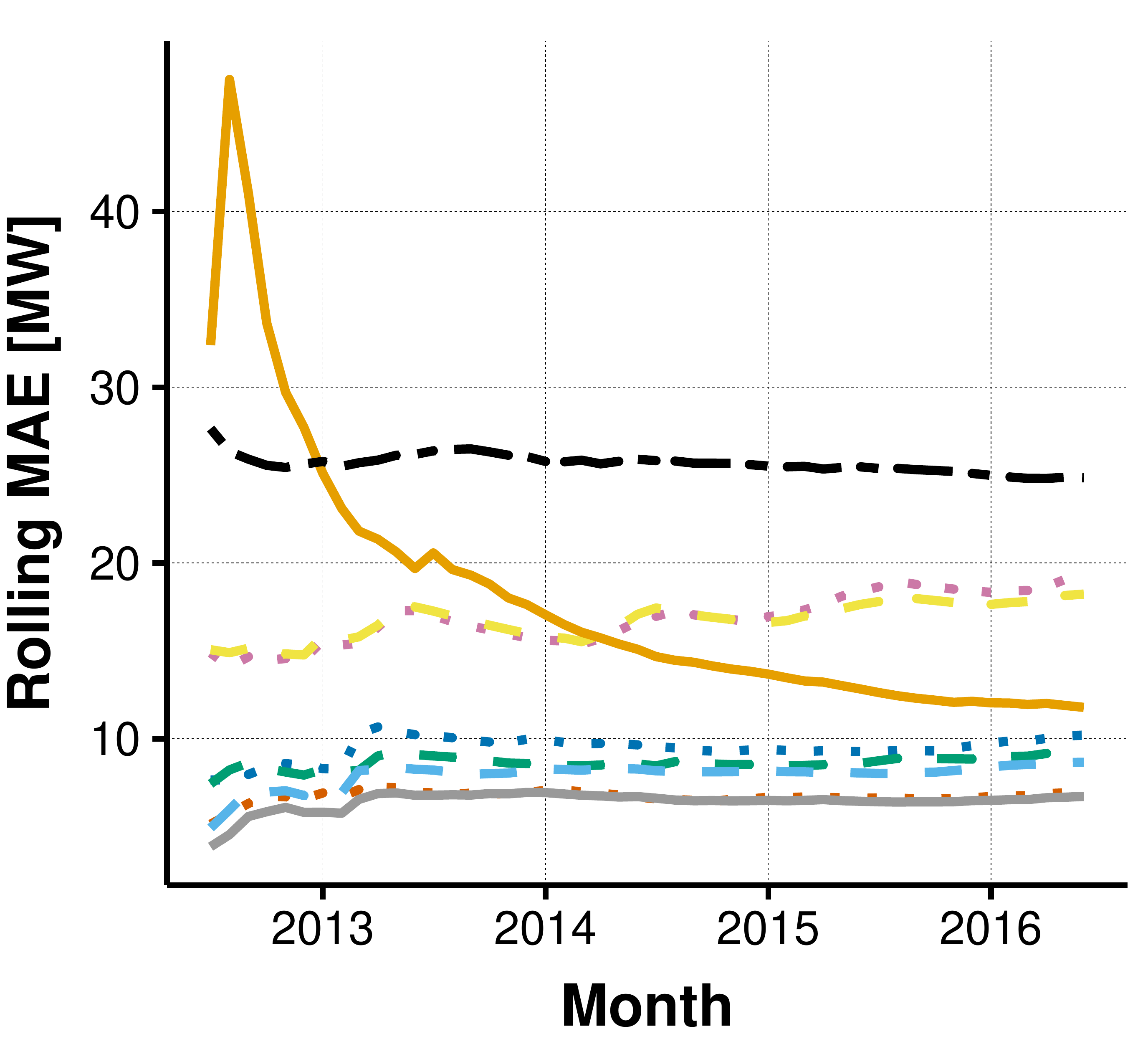}  \\
        (a) & (b)  \\[6pt]
        \includegraphics[width=\linewidth,keepaspectratio]{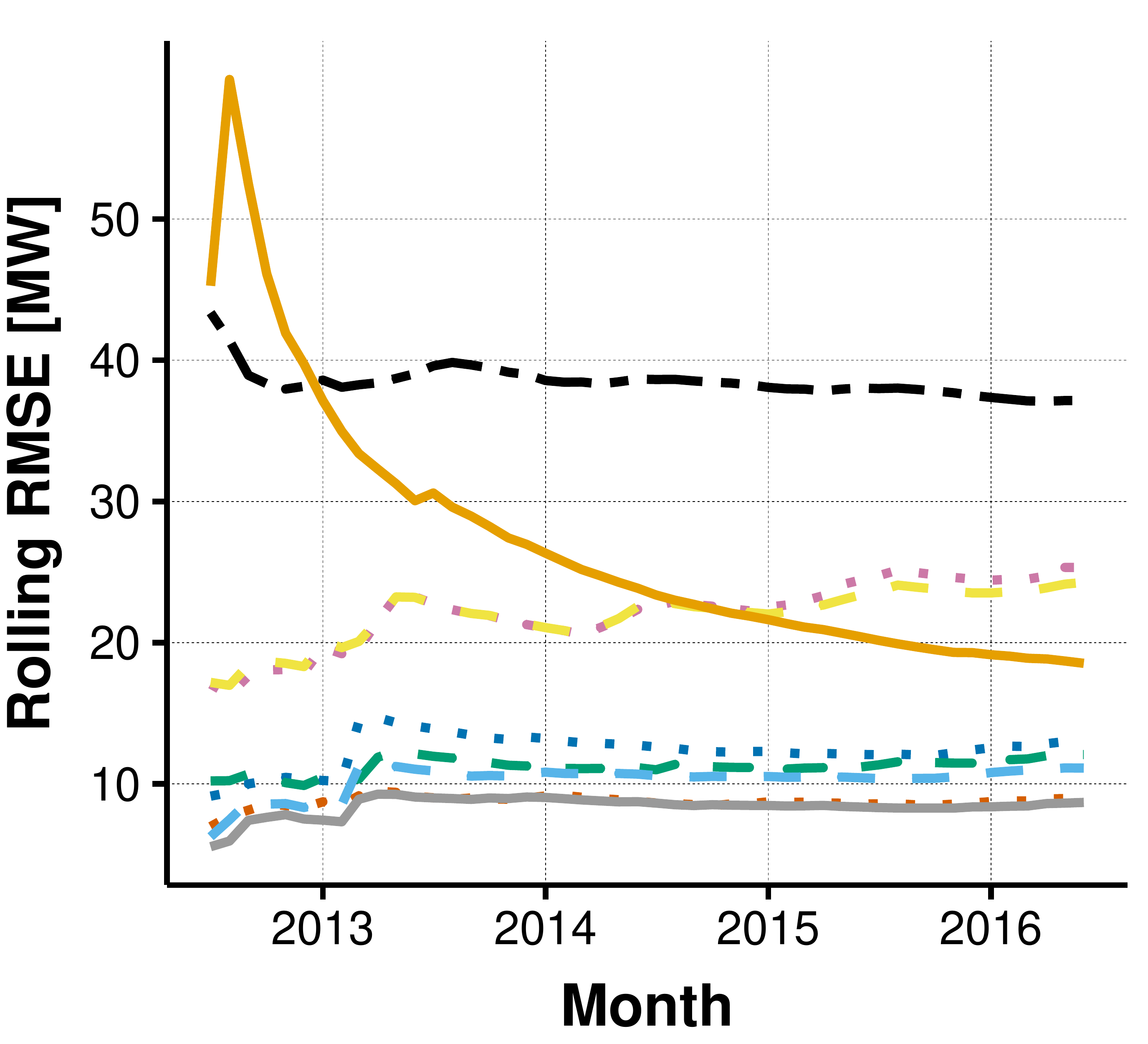} &
        \includegraphics[width=\linewidth,keepaspectratio,page=2]{legend-DP.pdf} \\
        (c) & \\[6pt]
    \end{tabular}
\caption{Cumulative forecasting metrics evolution for each of the monthly updated DP models: (a) MAPE, (b) MAE, (c) RMSE}
\label{cumulative_DP}
\end{figure}

\begin{figure}[H]
    \centering
    \begin{tabular}{>{\centering\arraybackslash}m{0.45\linewidth} >{\centering\arraybackslash}m{0.45\linewidth} }
        \includegraphics[width=\linewidth,keepaspectratio]{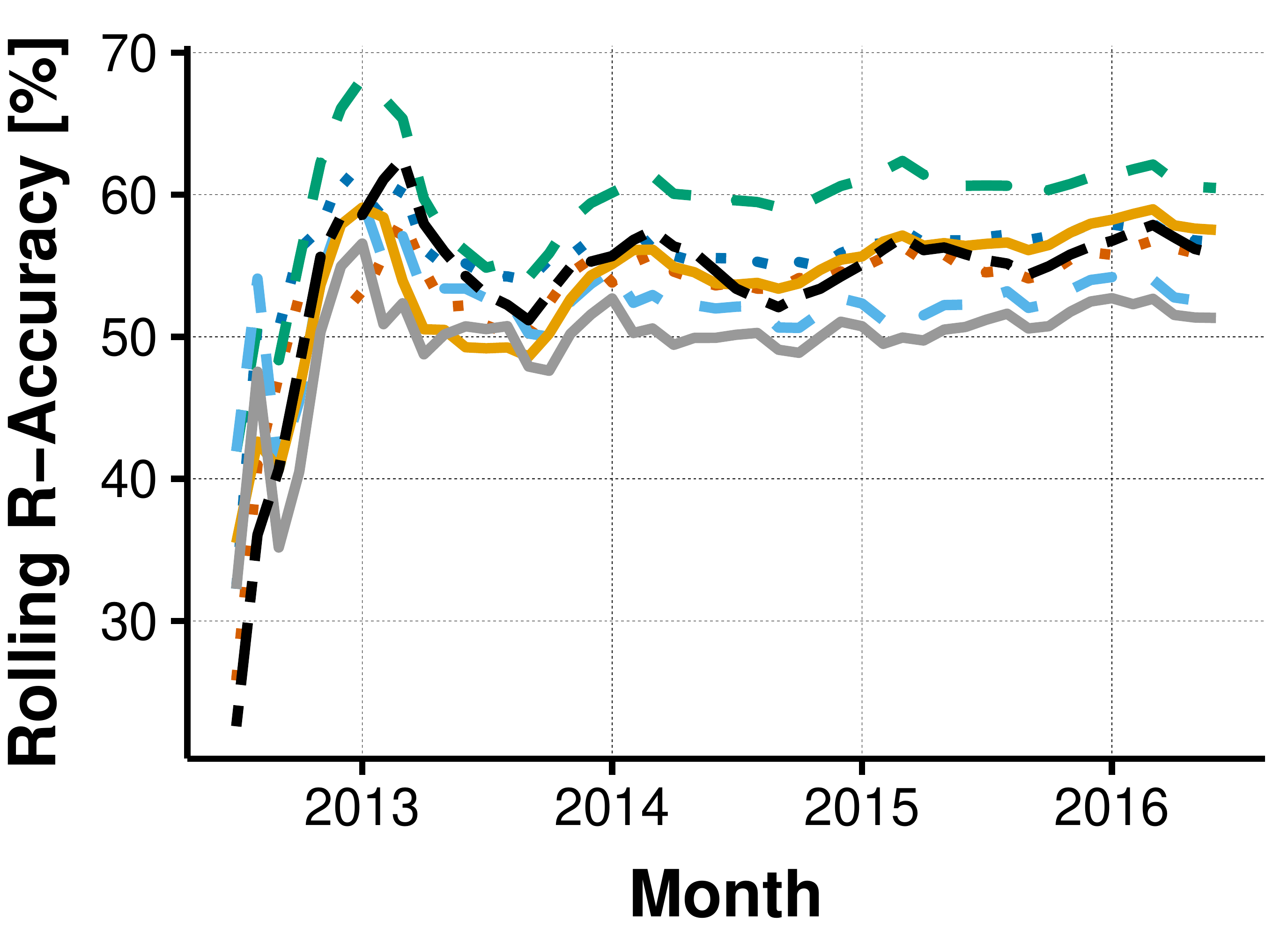} &
        \includegraphics[width=\linewidth,keepaspectratio]{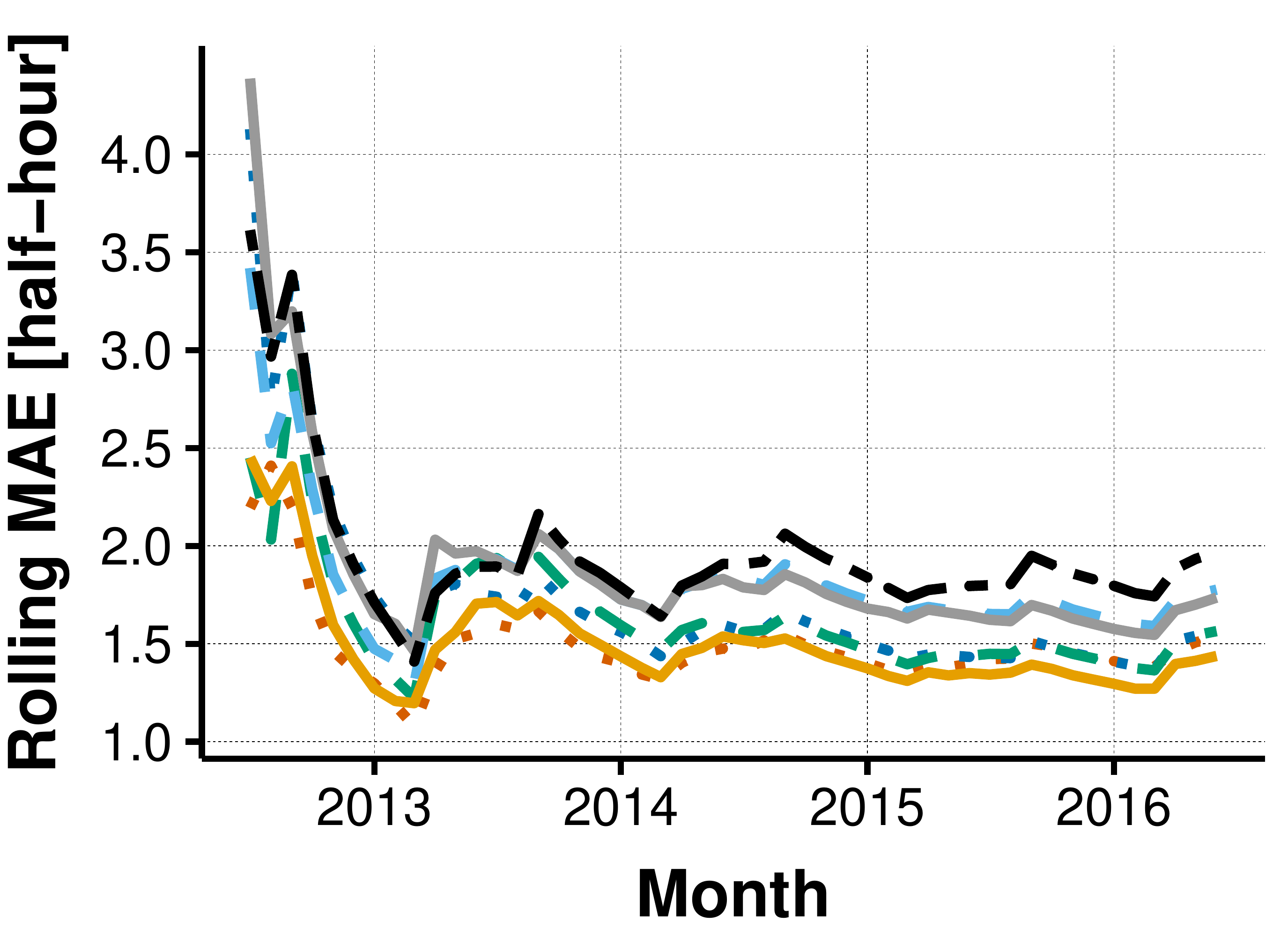}  \\
        (a) & (b)  \\[6pt]
        \includegraphics[width=\linewidth,keepaspectratio]{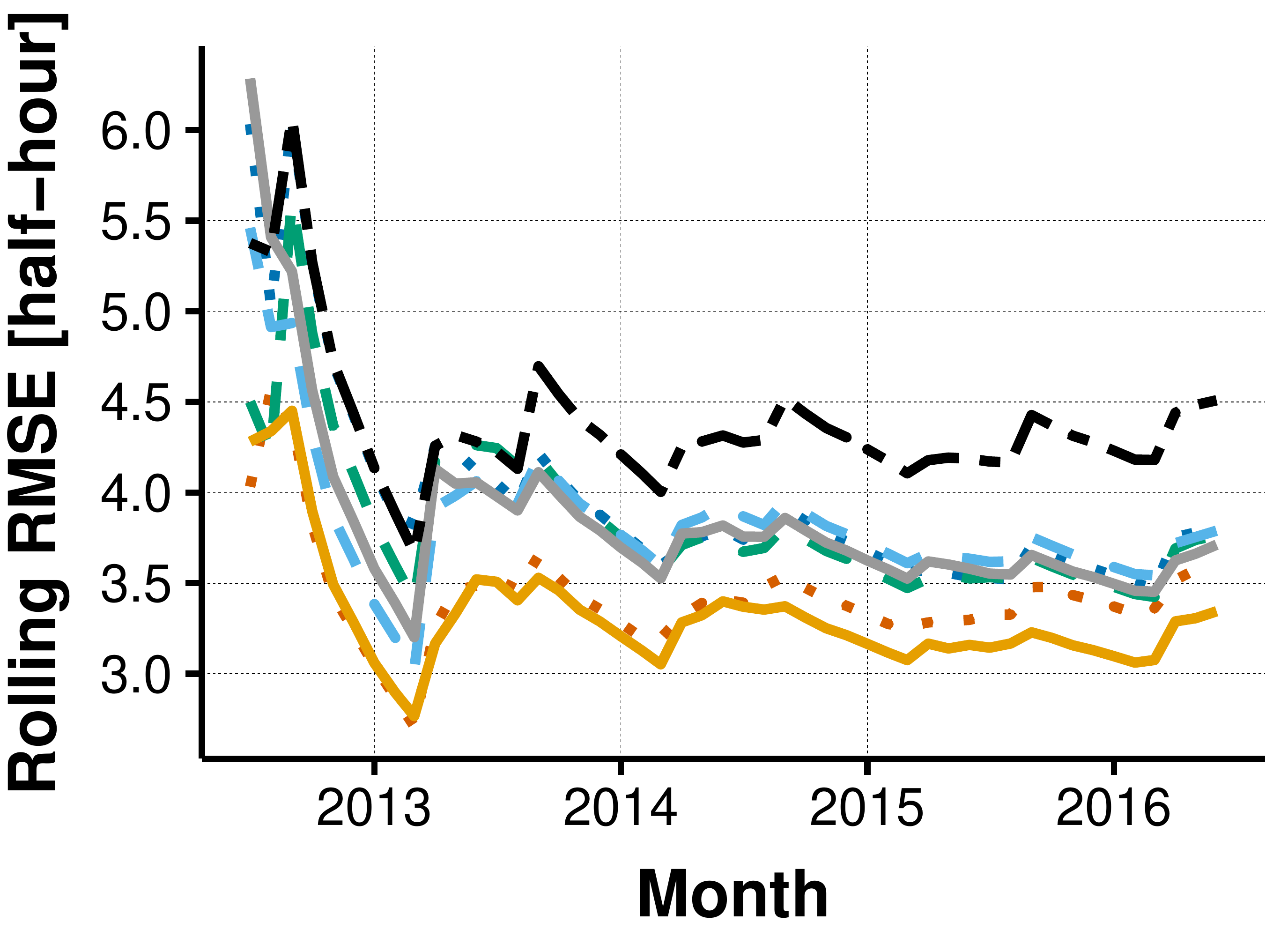} &
        \includegraphics[width=\linewidth,keepaspectratio,page=2]{legend-IP.pdf} \\
        (c) & \\[6pt]
    \end{tabular}
\caption{Cumulative forecasting metrics evolution for each of the monthly updated IP models: (d) R-accuracy, (e) MAE, (f) RMSE}
\label{cumulative_IP}
\end{figure}

As mentioned above, the rolling-origin forecasting setting may present a transitional regime during the first few training iterations. Figure \ref{cumulative_DP} and \ref{cumulative_IP} show the evolution of the different cumulative metrics calculated on the prediction signal updated on a monthly basis. Interestingly, the multi-resolution CNN for the DP (Figure \ref{cumulative_DP}) starts off with a very bad prediction error on the first months. With more data, its performance rapidly improves across all metrics. The other models have a less dramatic performance trend, with the multi-resolution GAMs consistently performing better than the other models. The prediction error of these models oscillates during the first few months, which can be explained by the fact that the models did not have enough information to adequately estimate the yearly cycle, because they were fitted to only one year of data. After a year, the prediction errors has stabilised. 

\begin{figure}[H] 
    \centering
    \includegraphics[width=\linewidth,keepaspectratio]{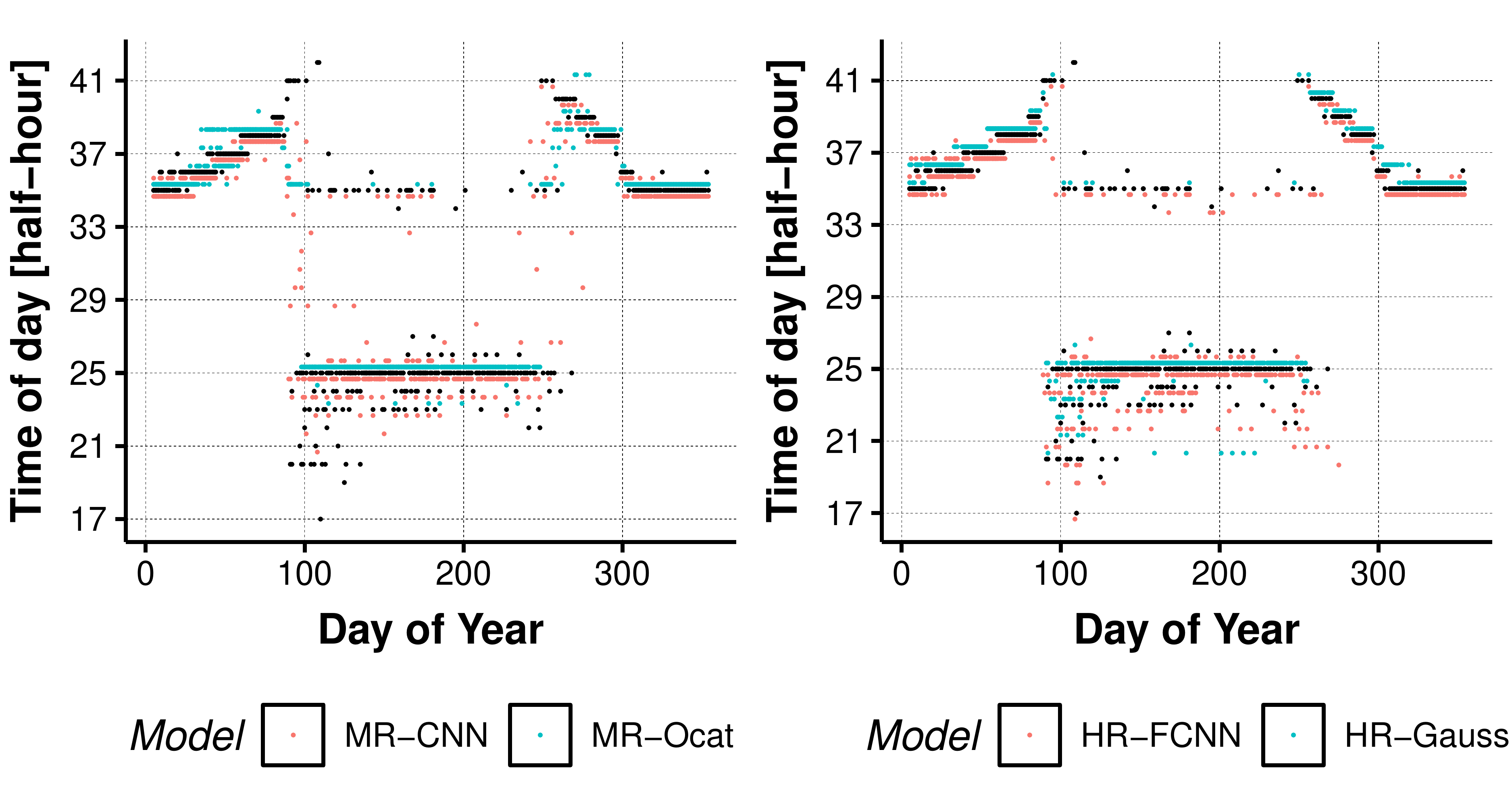}
    \caption{Left: observed IP as a function of the day of year (black) and corresponding predictions from MR-CNN (red, shifted downward for visibility) and MR-ocat (blue, shifted upward). Right: same plot for HR-FCNN (red, downward) and HR-Gauss (blue, upward).}
    \label{fig:ocat_issue_1}
\end{figure}

\begin{figure}[H] 
    \centering
    \begin{tabular}{>{\centering\arraybackslash}m{0.45\linewidth} >{\centering\arraybackslash}m{0.45\linewidth} }
        \includegraphics[width=\linewidth,keepaspectratio,page=8]{IP-block-bootstrap.pdf} &
        \includegraphics[width=\linewidth,keepaspectratio]{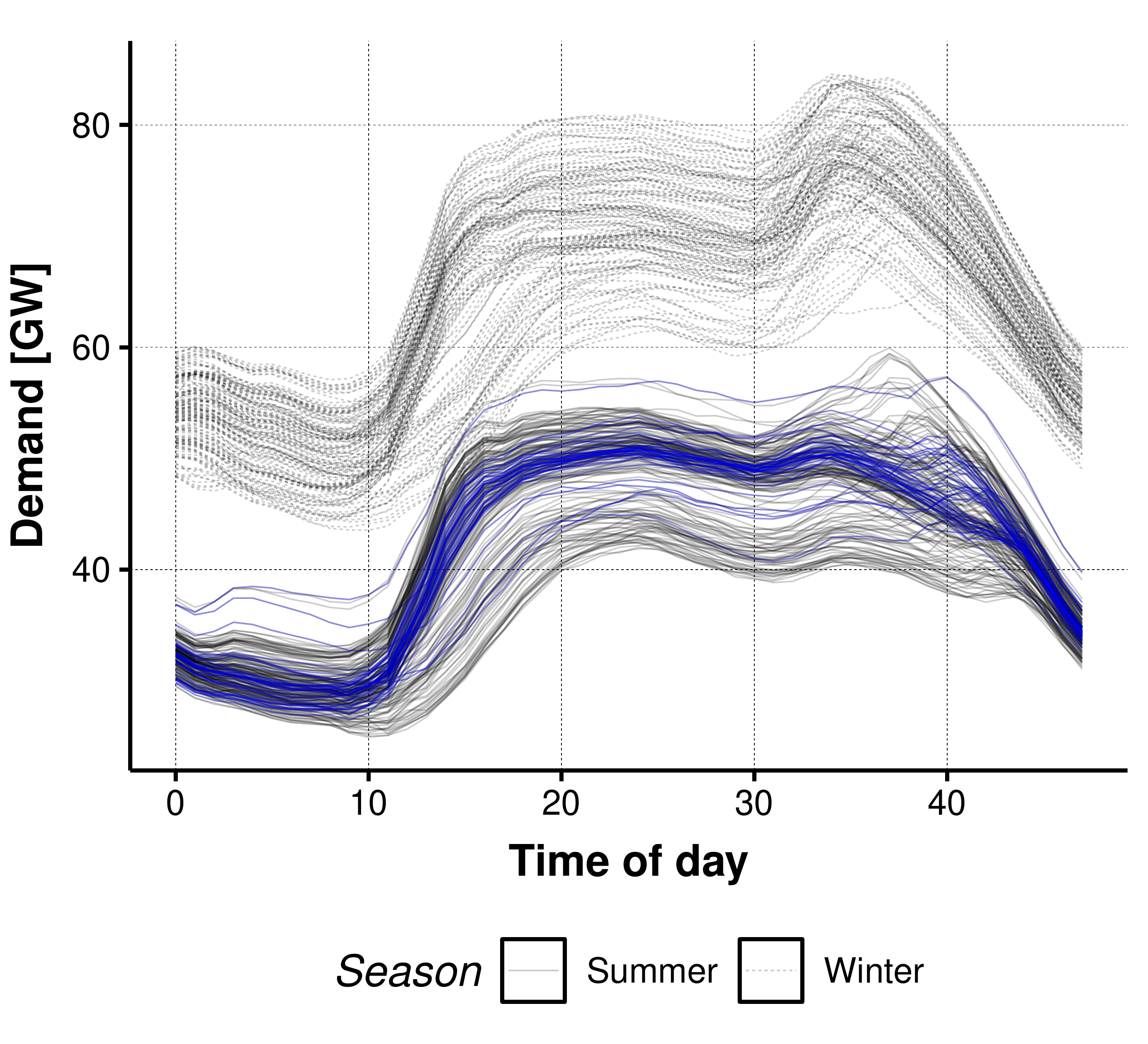}  \\
    \end{tabular}
    \caption{Left: Block-bootstrap boxplots of the d-RMSE metric for the IP problem. Right: daily demand profile curves during winter (shifted upward by 15 GW) and summer. The blue curves are profiles with a small absolute difference between the  morning and evening peak ($<$ 50 MW).}
    \label{fig:ocat_issue_2}
\end{figure}

For the IP forecasting task, the different metrics evolve with similar patterns (Figure \ref{cumulative_IP}), but the seasonal oscillations in performance persist beyond the first year. Figure \ref{fig:ocat_issue_1} explains why predicting the IP is harder in summer than in winter. In particular, while winter daily demand profiles have a reliable evening peak, summer load profiles are flatter and on some days the peak distribution becomes bimodal. That is, the daily peak might occur in the morning and or in the evening with equal probability. This is shown also by the right plot in Figure \ref{fig:ocat_issue_2}. Hence, it is clear that in the summer the IP point estimates might be unfairly penalised under the simple metrics considered here. This implies that a forecasting model might be better off providing an IP forecast that falls between the two peaks, as MR-CNN is occasionally doing (see Figure \ref{fig:ocat_issue_1}). Such a forecast might improve the metrics but has little value in an operational setting. Note also that the ocat model struggles to capture an IP distribution that is unimodal or bimodal depending on the time of year. In particular, the ocat model used here is based on a standard ordered-logit parametrisation, which involves modelling the mean of a latent logistic random variable via an additive model. It is not possible to transform a unimodal distribution on the ordered categories (here, IP) into a bimodal one, simply by controlling a location parameter. Hence, a more flexible model (e.g., \citealp*{peterson1990partial}) would be preferable.

It is interesting to verify the performance of each model for IP forecasting via a bespoke metric. In particular, let $t^{\text{m}}_i$ be the observed IP on day $i$ and let $\hat{t}^{\text{m}}_i$ be the corresponding forecast. We propose the following metric:
\begin{equation*}
\text{d-RMSE} = \left(\frac{1}{n}\sum_{i=1}^n(y_{t^{\text{m}}_i} - y_{\hat{t}^{\text{m}}_i})\right)^{1/2}
\end{equation*}
which is based on the difference between the daily peak demand and the demand at the predicted IP (the d stands for demand). This metric is more relevant to operations than MSE or MAE. For instance, in peak shaving applications, providing a forecast $\hat{t}^{\text{m}}_i$ very different from $t^{\text{m}}_i$ might not be a problem if $y_{t^{\text{m}}_i}$ and $y_{\hat{t}^{\text{m}}_i}$ are similar, which is what d-RMSE quantifies. Figure \ref{fig:ocat_issue_2} shows a bootstrapped boxplot of d-RMSE for each model. Interestingly, high-resolution methods are best here, by a substantial margin in the case of HR-FCNN.

The results obtained so far do not provide reliable evidence in favour or against the adoption of a multi-resolution approach for IP forecasting. In fact, the poor forecasting performance of MR-ocat is arguably attributable to the particular ordered-logit parametrisation used here. MR-CNN does well using standard, statistically motivated losses but it is inferior to high-resolution approaches on an operationally relevant one (d-RMSE). It would be interesting to verify whether fitting the MR-CNN model by minimising d-RMSE directly (rather than MSE as done here) would lead to better results. We leave this, and the search for a more flexible distribution for ordered categorical responses, for future work.

Implementing the multi-resolution approach on the DP forecasting problem is more straightforward, hence the results discussed so far are positive and reliable. We further verify their significance by performing \cite*{diebold_comparing_1995} (DM) tests on the absolute and squared error losses . The null hypothesis of the tests is: ``both forecasts have the same expected loss''. The results of the DM tests are available on Figure \ref{DMTEST} which confirms that, within the GAM class, the multi-resolution forecasts are significantly different to the low-resolution and high-resolution approaches under both metrics. 

\begin{figure}[H]
    \centering
    \begin{tabular}{c}
        \resizebox{\textwidth}{!}{%
        \begin{tabular}{lllllllllllll}
Model    & HR-arima                 & HR-gauss                 & HR-FCNN                  & LR-arima                 & LR-gauss                     & LR-scat                  & LR-gev                   & LR-FCNN                      & MR-gauss                   & MR-scat                    & MR-gev                     & MR-CNN                       \\
HR-arima & \cellcolor[HTML]{C0C0C0} & 0                        & 0                        & \textcolor{red}{0.116}                    & 0                            & 0                        & 0 & 0                            & 0                          & 0                          & 0                          & 0                            \\
HR-gauss & \cellcolor[HTML]{C0C0C0} & \cellcolor[HTML]{C0C0C0} & 0                        & 0                        &  0.042 & 0                        & 0 & 0.001 & 0                          & 0                          & 0                          & 0                            \\
HR-FCNN  & \cellcolor[HTML]{C0C0C0} & \cellcolor[HTML]{C0C0C0} & \cellcolor[HTML]{C0C0C0} & 0                        & 0                            & 0                        & 0 & 0     & \textcolor{red}{0.729}                      & \textcolor{red}{0.549}                      & \textcolor{red}{0.250}                      &  0.029\\
LR-arima & \cellcolor[HTML]{C0C0C0} & \cellcolor[HTML]{C0C0C0} & \cellcolor[HTML]{C0C0C0} & \cellcolor[HTML]{C0C0C0} & 0                            & 0                        & 0 & 0     & 0                          & 0                          & 0                          & 0     \\
LR-gauss & \cellcolor[HTML]{C0C0C0} & \cellcolor[HTML]{C0C0C0} & \cellcolor[HTML]{C0C0C0} & \cellcolor[HTML]{C0C0C0} & \cellcolor[HTML]{C0C0C0}     & 0                        & 0 & { 0.010} & 0                          & 0                          & 0                          & 0     \\
LR-scat  & \cellcolor[HTML]{C0C0C0} & \cellcolor[HTML]{C0C0C0} & \cellcolor[HTML]{C0C0C0} & \cellcolor[HTML]{C0C0C0} & \cellcolor[HTML]{C0C0C0}     & \cellcolor[HTML]{C0C0C0} & 0 & \textcolor{red}{0.122}                             & 0                          & 0                          & 0                          & 0     \\
LR-gev   & \cellcolor[HTML]{C0C0C0} & \cellcolor[HTML]{C0C0C0} & \cellcolor[HTML]{C0C0C0} & \cellcolor[HTML]{C0C0C0} & \cellcolor[HTML]{C0C0C0}     & \cellcolor[HTML]{C0C0C0} & \cellcolor[HTML]{C0C0C0} & 0                            & 0                          & 0                          & 0                          & 0     \\
LR-FCNN  & \cellcolor[HTML]{C0C0C0} & \cellcolor[HTML]{C0C0C0} & \cellcolor[HTML]{C0C0C0} & \cellcolor[HTML]{C0C0C0} & \cellcolor[HTML]{C0C0C0}     & \cellcolor[HTML]{C0C0C0} & \cellcolor[HTML]{C0C0C0} & \cellcolor[HTML]{C0C0C0}     & 0                          & 0                          & 0                          & 0     \\
MR-gauss & \cellcolor[HTML]{C0C0C0} & \cellcolor[HTML]{C0C0C0} & \cellcolor[HTML]{C0C0C0} & \cellcolor[HTML]{C0C0C0} & \cellcolor[HTML]{C0C0C0}     & \cellcolor[HTML]{C0C0C0} & \cellcolor[HTML]{C0C0C0} & \cellcolor[HTML]{C0C0C0}     & \cellcolor[HTML]{C0C0C0} & \textcolor{red}{0.063}                         & 0                          & { 0.001} \\
MR-scat  & \cellcolor[HTML]{C0C0C0} & \cellcolor[HTML]{C0C0C0} & \cellcolor[HTML]{C0C0C0} & \cellcolor[HTML]{C0C0C0} & \cellcolor[HTML]{C0C0C0}     & \cellcolor[HTML]{C0C0C0} & \cellcolor[HTML]{C0C0C0} & \cellcolor[HTML]{C0C0C0}     & \cellcolor[HTML]{C0C0C0}   & \cellcolor[HTML]{C0C0C0} & 0                          & 0                            \\
MR-gev   & \cellcolor[HTML]{C0C0C0} & \cellcolor[HTML]{C0C0C0} & \cellcolor[HTML]{C0C0C0} & \cellcolor[HTML]{C0C0C0} & \cellcolor[HTML]{C0C0C0}     & \cellcolor[HTML]{C0C0C0} & \cellcolor[HTML]{C0C0C0} & \cellcolor[HTML]{C0C0C0}     & \cellcolor[HTML]{C0C0C0}   & \cellcolor[HTML]{C0C0C0}   & \cellcolor[HTML]{C0C0C0} & \textcolor{red}{0.143}                         \\
MR-CNN   & \cellcolor[HTML]{C0C0C0} & \cellcolor[HTML]{C0C0C0} & \cellcolor[HTML]{C0C0C0} & \cellcolor[HTML]{C0C0C0} & \cellcolor[HTML]{C0C0C0}     & \cellcolor[HTML]{C0C0C0} & \cellcolor[HTML]{C0C0C0} & \cellcolor[HTML]{C0C0C0}     & \cellcolor[HTML]{C0C0C0}   & \cellcolor[HTML]{C0C0C0}   & \cellcolor[HTML]{C0C0C0}   & \cellcolor[HTML]{C0C0C0}  
\end{tabular}} \\
        (a) \\ [6pt]
\resizebox{\textwidth}{!}{%
\begin{tabular}{lllllllllllll}
Model     & HR-arima                                        & HR-gauss                                        & HR-FCNN                                         & LR-arima                                        & LR-gauss                                        & LR-scat                                         & LR-gev                                          & LR-FCNN                                         & MR-gauss                                        & MR-scat                                         & MR-gev                                          & MR-CNN                                          \\
HR-arima  & \cellcolor[HTML]{C0C0C0}{ } & 0                        & 0                        & \textcolor{red}{0.161}                    & 0                        & 0                        & 0                        & 0                        & 0                        & 0                        & 0                        & 0                        \\
HR-gauss & \cellcolor[HTML]{C0C0C0}{ } & \cellcolor[HTML]{C0C0C0}{ } & 0                        & 0                        & { 0.010}                    & 0                        & { 0.003}                    & \textcolor{red}{0.118}                    & 0                        & 0                        & 0                        & 0                        \\
HR-FCNN    & \cellcolor[HTML]{C0C0C0}{ } & \cellcolor[HTML]{C0C0C0}{ } & \cellcolor[HTML]{C0C0C0}{ } & 0                        & 0                        & 0                        & 0                        & 0                        & \textcolor{red}{0.400}                    & \textcolor{red}{0.362}                    & \textcolor{red}{0.492}                    & { 0.016}                    \\
LR-arima & \cellcolor[HTML]{C0C0C0}{ } & \cellcolor[HTML]{C0C0C0}{ } & \cellcolor[HTML]{C0C0C0}{ } & \cellcolor[HTML]{C0C0C0}{ } & 0                        & 0                        & 0                        & 0                        & 0                        & 0                        & 0                        & 0                        \\
LR-gauss & \cellcolor[HTML]{C0C0C0}{ } & \cellcolor[HTML]{C0C0C0}{ } & \cellcolor[HTML]{C0C0C0}{ } & \cellcolor[HTML]{C0C0C0}{ } & \cellcolor[HTML]{C0C0C0}{ } & 0                        & 0                        & \textcolor{red}{0.938}                    & 0                        & 0                        & 0                        & 0                        \\
LR-scat  & \cellcolor[HTML]{C0C0C0}{ } & \cellcolor[HTML]{C0C0C0}{ } & \cellcolor[HTML]{C0C0C0}{ } & \cellcolor[HTML]{C0C0C0}{ } & \cellcolor[HTML]{C0C0C0}{ } & \cellcolor[HTML]{C0C0C0}{ } & 0                        & { 0.026}                    & 0                        & 0                        & 0                        & { 0.002}                    \\
LR-gev   & \cellcolor[HTML]{C0C0C0}{ } & \cellcolor[HTML]{C0C0C0}{ } & \cellcolor[HTML]{C0C0C0}{ } & \cellcolor[HTML]{C0C0C0}{ } & \cellcolor[HTML]{C0C0C0}{ } & \cellcolor[HTML]{C0C0C0}{ } & \cellcolor[HTML]{C0C0C0}{ } & 0                   & 0                        & 0                        & 0                        & 0                        \\
LR-FCNN    & \cellcolor[HTML]{C0C0C0}{ } & \cellcolor[HTML]{C0C0C0}{ } & \cellcolor[HTML]{C0C0C0}{ } & \cellcolor[HTML]{C0C0C0}{ } & \cellcolor[HTML]{C0C0C0}{ } & \cellcolor[HTML]{C0C0C0}{ } & \cellcolor[HTML]{C0C0C0}{ } & \cellcolor[HTML]{C0C0C0}{ } & 0                        & 0                        & 0                        & 0                        \\
MR-gauss & \cellcolor[HTML]{C0C0C0}{ } & \cellcolor[HTML]{C0C0C0}{ } & \cellcolor[HTML]{C0C0C0}{ } & \cellcolor[HTML]{C0C0C0}{ } & \cellcolor[HTML]{C0C0C0}{ } & \cellcolor[HTML]{C0C0C0}{ } & \cellcolor[HTML]{C0C0C0}{ } & \cellcolor[HTML]{C0C0C0}{ } & \cellcolor[HTML]{C0C0C0}{ } & \textcolor{red}{0.468}                    & 0                        & 0                        \\
MR-scat  & \cellcolor[HTML]{C0C0C0}{ } & \cellcolor[HTML]{C0C0C0}{ } & \cellcolor[HTML]{C0C0C0}{ } & \cellcolor[HTML]{C0C0C0}{ } & \cellcolor[HTML]{C0C0C0}{ } & \cellcolor[HTML]{C0C0C0}{ } & \cellcolor[HTML]{C0C0C0}{ } & \cellcolor[HTML]{C0C0C0}{ } & \cellcolor[HTML]{C0C0C0}{ } & \cellcolor[HTML]{C0C0C0}{ } & 0                        & 0                        \\
MR-gev   & \cellcolor[HTML]{C0C0C0}{ } & \cellcolor[HTML]{C0C0C0}{ } & \cellcolor[HTML]{C0C0C0}{ } & \cellcolor[HTML]{C0C0C0}{ } & \cellcolor[HTML]{C0C0C0}{ } & \cellcolor[HTML]{C0C0C0}{ } & \cellcolor[HTML]{C0C0C0}{ } & \cellcolor[HTML]{C0C0C0}{ } & \cellcolor[HTML]{C0C0C0}{ } & \cellcolor[HTML]{C0C0C0}{ } & \cellcolor[HTML]{C0C0C0}{ } & { 0.015}                    \\
MR-CNN     & \cellcolor[HTML]{C0C0C0}{ } & \cellcolor[HTML]{C0C0C0}{ } & \cellcolor[HTML]{C0C0C0}{ } & \cellcolor[HTML]{C0C0C0}{ } & \cellcolor[HTML]{C0C0C0}{ } & \cellcolor[HTML]{C0C0C0}{ } & \cellcolor[HTML]{C0C0C0}{ } & \cellcolor[HTML]{C0C0C0}{ } & \cellcolor[HTML]{C0C0C0}{ } & \cellcolor[HTML]{C0C0C0}{ } & \cellcolor[HTML]{C0C0C0}{ } & \cellcolor[HTML]{C0C0C0}{ }
\end{tabular}} \\
        (b) \\[6pt]

    \end{tabular}
\caption{P-values from the Diebold-Mariano test for DP forecasts. The test used is from the \textit{multDM} package in R \citep*{drachal_multdm_2020}. In black, the null hypothesis is rejected at the 5\% threshold and both forecasts are significantly different. In red, the null hypothesis is not rejected at the 5\% threshold and both forecasts cannot be significantly differentiated; (a) absolute errors (b) squared errors.}
\label{DMTEST}
\end{figure}

\begin{figure}[H]
    \centering
    \includegraphics[width=0.8\linewidth,keepaspectratio]{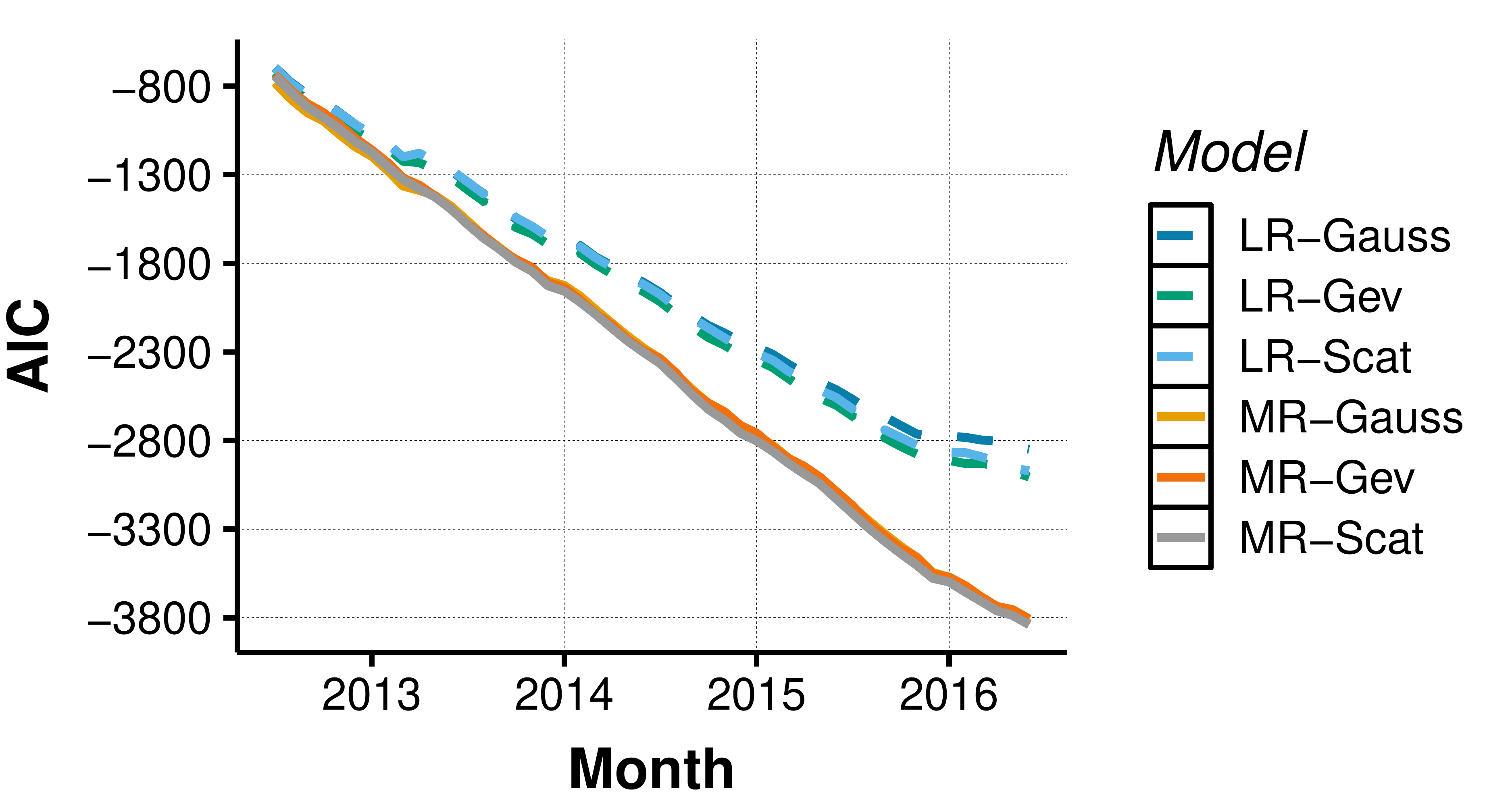}
    \caption{AIC for the low-resolution and multi-resolution DP GAMs}
    \label{AIC}
\end{figure}

It is interesting to quantify the complexity or parsimony of the models considered so far. AIC can be interpreted as a parsimony measure, but it requires computing the effective number of models parameters and we are not aware of any method that would allow estimating them across all the model classes considered here. Figure \ref{AIC} shows the AICs of low- and multi-resolution GAMs. The multi-resolution approaches consistently have a smaller AIC than the low-resolution approaches. Furthermore, the slopes indicate that with more data the gap continues to increase. 

For NNs, parsimony is highly dependent on the chosen architecture. In our case, the low-resolution and high-resolution NNs have a very similar architecture with only one hidden layer and a dropout layer (Figure \ref{HRFCNN} and Figure \ref{LRFCNN}). Only the input shapes and the number of observations vary. On the other hand, the multi-resolution NN (Figure \ref{MRCNN}) requires the use of convolutional layers which are leveraged to extract the high-resolution information. The extraction process requires multiple layers which forces the multi-resolution CNN to have a larger number of parameters than the low-resolution and high-resolution NNs.

The results discussed in this section show that multi-resolution approaches are superior to low- and high-resolution alternatives for the DP forecasting problem. The forecasting performance of the high-resolution FCNN and the multi-resolution GAMs are not significantly different but, in an operational peak demand forecasting context, the multi-resolution GAM would be preferred because it can be decomposed into additive components, which can be more easily interpreted (and manually adjusted) by operational staff. In addition, note that adopting a multi-resolution approach can bring substantial computational advantages, which are easy to quantify within the GAM model class. In particular, the GAM model matrix $\bf X$ in the multi-resolution case has $T$ times less rows than in the high-resolution case, where $T$ is the number of daily observations (i.e., $T=48$ for half-hourly data). Therefore, $T$ times less memory is used, and many computations frequently required during GAM model fitting (such as ${\bf X}^T {\bf W} {\bf X}$, where $\bf W$ is a diagonal matrix) will take less time. 

\section{Conclusion}

This paper proposes a novel modelling approach, which uses both high-resolution and low-resolution information to forecast the daily electrical load peak magnitude and timing. The results demonstrate that this multi-resolution approach is flexible enough to be applied to different model classes and that it provides a competitive predictive performance. In particular, GAMs and NNs with similar input structures were used to implement the multi-resolution approach and to compare its performance that of low-resolution, high-resolution and persistence alternatives. On UK aggregate demand data, the multi-resolution models performed significantly better across all metrics when forecasting peak magnitude. In addition to improved predictions, adopting a multi-resolution approach enables faster computation via data compression and leads to more parsimonious models, as demonstrated by the consistently lower AIC scores achieved by multi-resolution models within the GAM model class. 

The results on the peak timing forecasting problem are mixed, but interesting. A multi-resolution neural network does marginally better than the alternatives, when performance is assessed via standard statistical metrics. However, the corresponding forecast is occasionally inappropriate (falling between the morning and evening peaks) and inferior to high-resolution alternatives when assessed via an operationally motivated metric. The results suggest that the multi-resolution neural network should be fitted to data by minimising a problem specific performance metric directly. For instance, one could consider financial metrics on billing periods as done by \cite*{saxena_hybrid_2019}. The multi-resolution GAM does poorly on the peak timing problem, but this is attributable to the insufficient flexibility of the ordered logit parametrisation used here. Obtaining stronger evidence in favour or against the use of multi-resolution methods for the peak timing problem would require solving the issues just mentioned, which could be the subject of further work.

The forecasting methods presented here could be extended in several ways. The set of models described in this paper could be used within an aggregation of experts or ensemble methods, which might lead to more accurate forecasts. The benefits of multi-resolution methods have been demonstrated in a context where covariates were available at different temporal resolutions, but they could be generalised to other multi-resolution settings, such as spatio-temporal data or individual customer data (see e.g., \citealp*{fasiolo_qgam_2020} for an example application of functional quantile GAMs \citealp*{fasiolo_fast_2020} to residential electricity demand data). Finally, this paper focused on day-ahead daily peak magnitude and time forecasting, but multi-resolution methods could be applied to other short-term windows (e.g., weekly). However,  estimating monthly or yearly peaks would require a different approach, because the number of observed demand peaks would be too low.

\section*{Acknowledgments}
Matteo Fasiolo was partially funded by EPSRC grant EP/N509619/1. The datasets used in this paper are available on the National Grid and National Oceanic and Atmospheric Administration websites. The R code as well as the data prepared for the experiments in this paper are available at the following link: \href{https://cutt.ly/CYvgIP3}{https://cutt.ly/CYvgIP3}

\bibliography{references.bib}

\end{document}